%% file: template.tex
\documentclass{article}
\usepackage{arxiv}
\usepackage{hyperref}
\hypersetup{
  pdfinfo={
    Title={Regenerating Arbitrary Video Sequences with Distillation Path-Finding},
  }
}

\usepackage[utf8]{inputenc} 
\usepackage[T1]{fontenc}    
\usepackage{hyperref}       
\usepackage{url}            
\usepackage{booktabs}       
\usepackage{amsfonts}       
\usepackage{nicefrac}       
\usepackage{microtype}      
\usepackage{lipsum}
\usepackage{graphicx}
\graphicspath{ {./images/} }
\usepackage{amsmath}
\usepackage[numbers]{natbib}
\usepackage{dirtytalk}
\usepackage{subfig}
\usepackage{xcolor,soul,framed} 
\usepackage{enumitem}
\usepackage{algorithm} 
\usepackage{wrapfig}

\usepackage{array}
\usepackage{float}
\usepackage{mdwmath}
\usepackage{amsmath}
\usepackage{algpseudocode}
\usepackage{mdwtab}
\usepackage{eqparbox}
\usepackage{url}
\usepackage{csvsimple}
\usepackage{amssymb}
\usepackage{multicol}
\usepackage{multirow, tabularx}
\usepackage{adjustbox}
\usepackage{makecell}
\usepackage{siunitx}

\title{Regenerating Arbitrary Video Sequences with Distillation Path-Finding}

\author{
 Thi-Ngoc-Hanh Le\\
 National Cheng-Kung University\\
  Taiwan\\
  \texttt{ngochanh.le1987@gmail.com}
   \And
 Sheng-Yi Yao \\
 National Cheng-Kung University\\
  Taiwan\\
  \texttt{nd8081018@gs.ncku.edu.tw}
  \And
 Chun-Te Wu \\
 National Cheng-Kung University\\
  Taiwan\\
  \texttt{oojimda3838@hotmail.com}
  \And
  Tong-Yee Lee*\\  
  National Cheng-Kung University\\
  Taiwan\\
  \texttt{tonylee@mail.ncku.edu.tw}
}

\begin{document}
\maketitle

\begin{abstract}
If the video has long been mentioned as a widespread visualization form, the animation sequence in the video is mentioned as storytelling for people. Producing an animation requires intensive human labor from skilled professional artists to obtain plausible animation in both content and motion direction, incredibly for animations with complex content, multiple moving objects, and dense movement. This paper presents an interactive framework to generate new sequences according to the users' preference on the starting frame. The critical contrast of our approach versus prior work and existing commercial applications is that novel sequences with arbitrary starting frame are produced by our system with a consistent degree in both content and motion direction. To achieve this effectively, we first learn the feature correlation on the frameset of the given video through a proposed network called RSFNet. Then, we develop a novel path-finding algorithm, SDPF, which formulates the knowledge of motion directions of the source video to estimate the smooth and plausible sequences. The extensive experiments show that our framework can produce new animations on the cartoon and natural scenes and advance prior works and commercial applications to enable users to obtain more predictable results.


\end{abstract}

\section{Introduction}
Video has long been a widespread media form in our daily life. In addition to visualizing, the sequence of animation in a video is mentioned as storytelling for people. Animating production is usually a specialized and time-consuming job, requiring intensive human labor from skilled professional artists. In traditional cartoon animation (\textit{i.e., }cel-based and path-based animation) the procedure is complicated and needs much repeated manual labor, and a large amount of cartoon materials have been produced during this procedure. If all these material can be effectively managed and reused, we not only can speed up the time of producing an art but also easily create variations of the existing material. Although the recent computer-aid techniques have removed the burden of artists from tedious work in producing new animations, understanding the content (\textit{i.e.}, character’s gesture, background scene, etc.) and finding smooth transitions, are still challenging. The existing commercial applications, \textit{e.g.}, Toon Boom, Adobe Animate, mostly serve the capability on cartoon characters with basic animations. They lack diversity in animation styles and cartoon scenes. Therefore, it's necessary to establish and develop a cartoon images management and retrieval system supporting interactive fast animation making, so that the artists can pay more attention to the creative work, rather than those repeated work like colorizing, repainting, etc.

This problem has been explored. Previous work on this domain can be divided into feature-based and sequence-estimation methods. In feature-based methods, research attempts have been made to get knowledge on image content \citep{1_fried2017patch2vec, yu2012combining, 3_yang2010recognizing, de2004cartoon}. \citet{1_fried2017patch2vec} train a convolutional neural network to map images into lower dimensional space and define their similarity by a distance calculation. \citet{2_yu2014semantic} propose an algorithm to construct the feature space according to the shape context of the character in the image and the user's label. However, their dataset is labeled by human judgment, which is difficult or time-consuming to collect. Then, after projecting images into the feature space, the distance metric between two images can be considered as the similarity distance. Nevertheless, the user still needs to manually label the relation between the data. \citet{3_yang2010recognizing} extract three different features of the character in the image's shape context, color histogram, and motion direction. These features are then fused to result in the feature vectors of the character images. But, the segmentation of the character images required by their algorithm is not easy to generate correctly without professional skill.

In contrast, the sequence-estimation methods investigate different approaches to generate a plausible animation. \citet{5_schodl2000video} train a binary classifier and apply the Q-learning algorithm \citep{6_kaelbling1996reinforcement} on the images library to produce arbitrary length video sequences. \citet{7_yu2011semi} use a semisupervised algorithm to select the next frame of the initial frame according to the similarity distance. Then, they will treat the next frame as the initial frame and repeat this iterative process to generate the results. Recently, \citet{morace2022learning} construct a graph by the similarity distance of images and compute the shortest Hamiltonian path for \textcolor{black}{reconstructing the sequence from a set of un-ordered images.}

However, there are three major drawbacks in the above research. First, they solely focus on cartoon characters. Second, feature extraction and the distance metric used to measure such features are developed independently. And third, with these two mentioned issues, such a prior system is not sufficient to challenge the input clip that consists of dense motion and content. Therefore, we address the demanding problem by combining knowledge learned from a self-trained network and modeling them in a path-finding strategy to produce plausible and smooth videos efficiently.

In this paper, we propose a framework to address the above challenges. \textcolor{black}{We aim to create new smooth sequences according to users' preferences of the starting frame. We do not know the sequence we are to generate except the starting frame. Our designed framework attempts to minimize the artifacts caused by \textit{cold transition} and the \textit{flip-flop} phenomenon. The proposed framework pays attention to the pairwise relationship on both content and motion direction of an image and others in the image gallery. Our essential contribution to reducing user effort is automatically propagating user preference to predict a future sequence in a meaningful manner. To achieve this, we present a novel path-finding algorithm that absorbs the knowledge of features in our self-defined network and motion properties in the ground truth, which remedies the drawbacks of prior work.}

Our framework consists of an online knowledge learning and an offline sequence generation stage. The online stage learns the feature correlations of pairs of images in a given image set. These feature correlations serve as the initial guidance for new paths explored in the offline stage. The content of frames in real-world videos is complex in both background and foreground. Meanwhile, to model the user's selection to a plausible and novel animation, we need to calculate a meaningful degree of interchangeability between any two frames. We achieve this by proposing a neural network model, \textcolor{black}{\textit{Recursive-based Semantic Feature Network} (RSFNet)}, to learn the high-level representation of images. It is because the neighborhoods tend to be selected as correlation, which may prevent us from exploring new animations.

In the offline stage, the correlation of images learned in the online stage is performed in a graph. Users can specify their preferences for any node on the graph as the starting frame of their desired animation. Besides the meaningful degree, we need to preserve the temporal coherency in transitions. We tackle this by proposing an algorithm, \textit{Single-source Distillation Path-Finding} (SDPF), in which we embed constraints to interpret potential candidates for plausible animations. In summary, our main contributions are as follows:
\begin{itemize}
    \item A framework for resequencing videos, which exploits the feature correlation and the motion direction between frames to efficiently produce plausible and smooth video results.
    \item A framework to extract the representative feature vectors of the images in general style without requiring a large amount of dataset. And, the distance of the vectors can properly match the similarity of the images.
    \item A novel path-finding algorithm that can synthesize the resultant videos with smooth transitions from the image collection. Moreover, the random selection of our algorithm can increase the diversity of the results, and thus make each resultant sequence distinct from the others.
    \item Our overall system significantly reduces interaction time required to produce desired results. Besides, the proposed method works well in  both cartoon scenes and natural videos, and therefore this enables users to obtain more predictable results.
\end{itemize}

\section{Related work}
\subsection{\textcolor{black}{Feature Extraction and Dimension Reduction}}
\textcolor{black}{Researchers seek different approaches in analyzing images to learn the correlation between their representation. \citet{10_osadchy2007synergistic} propose an energy-based model to detect faces with different views. \citet{3_yang2010recognizing} use multiple features of cartoon characters to project images into lower dimensional space. \textcolor{black}{\citet{zhang2011excol} provide a flexible way for the extraction and completion steps to reflect the unique characteristics of cartoon animation}. The transductive algorithm \citep{4_gammerman2013learning} can fuse these different features together and construct a model which projects the character images into lower dimensional space. Combining multiple types of features \citep{2_yu2014semantic} has achieved great success in many areas. After extracting the feature vectors from the character’s shape in the image, users can provide image pairs' positive and negative relationships to restrict the distance between feature vectors. }

With the revolution of deep learning technologies, researchers develop the alternative promising approach. \citet{1_fried2017patch2vec} analyze patches by embedding them to a vector space in which the texture of image patches are considered to define the similarity of them. \citet{11_holden2015learning} use an autoencoder for human manifold. \citet{9_zhang2018local} propose an autoencoder architecture for image clustering. They first train the local stacked contractive autoencoder for the neighborhoods of training dataset based on Euclidean distance metric. \citet{14_zhang2019animation} use a convolutional autoencoder network to project the images into lower dimensional space, and the L2 distance between the latent vectors are considered as the similarity of the images. \citet{morace2022learning} utilize an off-the-shelf network LPIPS \citep{15_zhang2018unreasonable} to compute the similarity distance between images. \textcolor{black}{Most recently, \citet{xu2022multi} introduce a dual-task deep learning scheme for separating the structure content in a cartoon animation, i.e., content video and effect video.}

\textcolor{black}{Contrasting the above approaches, we handle arbitrary animation objects, including cartoon and natural scenes, rather than only focusing on cartoon characters. We get the knowledge of image representation by a self-defined network which is sufficient to capture comprehensive features. The network requires much less training dataset but has better performance than those in prior work. Plus, it enables us to be independent from such an intermediate network.}

\subsection{\textcolor{black}{Images sequence ordering}}
\textcolor{black}{Ordering a collection of images is usually considered as path-finding problem in a weighted graph, in which images are represented by vertices and the weights of the edges are the similarity of two end points, and other constraints such as temporal ordering, path smoothness, or user-control.}

\textcolor{black}{A variety of methods have been early developed to create sequences \citep{de2004cartoon, 5_schodl2000video, 13_schodl2002controlled}. Given a starting and ending frame, the system proposed in \citet{de2004cartoon} traverses on the manifold to re-sequence an existing cartoon library to a novel animation. Video textures \citep{5_schodl2000video} uses L2 distance of raw pixels of images as similarity and applies Q-learning algorithm \citep{6_kaelbling1996reinforcement} to generate an arbitrary length video sequence whose motion is similar to input video. Their method can produce convincing results in which input video has repetitive motion or unstructured stochastic motion. However, as the way to calculate similarity cannot adequately describe the high-level features of images, the case of complex structured motion such as human body motion will fail. To overcome this problem, \citet{13_schodl2002controlled} extract six specified features from the key-frames to train a binary classifier. This classifier will judge whether the transition between the key-frames will be accepted or not, and the cost of the transition depends on their original video sequences. They use the beam search to find the smoothest sequence. For the better result of the beam search, a hill climbing algorithm is used to interactively minimize the total cost of the sequence from initial random.}

\textcolor{black}{The seminal work has motivated researchers to investigate deeper recently \citep{7_yu2011semi, yu2012combining, 3_yang2010recognizing}. \citet{3_yang2010recognizing} present a cartoon gesture space to cartoon retrieval and synthesis. They use color, shape, and motion information in dissimilarity estimation. \citet{7_yu2011semi} propose a semi-supervised algorithm to create new cartoon animation from the image library. They extract the shape context of the characters in the images, and calculate the similarity distance based on the shape correspondence. Inspired by these methods, \citet{yu2012combining} use a semisupervised multiview subspace learning algorithm to encode different features in a unified space. To model the diverse dynamics, \citet{khan2021hamiltonian} introduce a deep generative model for image sequences, in which they split the motion space into subspaces and perform a unique Hamiltonian operator for each subspace.}

\textcolor{black}{Some different approaches are recently introduced \citep{14_zhang2019animation, morace2022learning}. To create a sequence, \citet{14_zhang2019animation} embed image collection into a convolutional autoencoder network. They then build the proximity graph based on the complete graph of the latent vectors and apply Monte Carlo algorithm to find the smoothest animation sequence. Meanwhile, \citet{morace2022learning} remove the last 10 percent outliers according to the generalized gamma probability distribution to fine-tune the smoothness of sequence. Then, they find the shortest Hamiltonian path to generate the resequencing results.}

\textcolor{black}{The sharp contrast between our framework and theirs \citep{de2004cartoon,yu2012combining, 3_yang2010recognizing, 14_zhang2019animation, morace2022learning} is that we develop a novel path-finding algorithm SDPF \textcolor{black}{to generate new sequences with arbitrary starting frame}. Our SDPF is faster than a greedy path-finding, effective to explore novel sequence and control the motion consistency. } 

\section{System overview}
The framework of our video resequencing is illustrated in Fig.\ref{fig_framework}, which consists of two primary models: a \textit{semantic relation graph} (SRG) model for representing the relation of images in the given set of images, and a \textit{Single-source Distillation path-finding} (SDPF) algorithm for exploring a path on SRG to resequence the video. Our system takes as input a video, we aim to generate new smooth sequences with arbitrary starting frame while maintaining the consistency in both content relations and temporal coherency.
\input{overall_framework}

The SRG models the set of frames in the given video to a completed graph. RSFNet explores this, i.e., the network we propose in this paper. RSFNet shoulders the task of converting images ($\{x_i\}$) into feature representation ($\{v_i\}$) in which every single $v_i$ represents a node in SRG. \textcolor{black}{To describe the semantic relation of $v_i$, we merge the triplet of recursive-based encoders (called R-Encoder) as a single one, \textit{i.e}., RSFNet, and train it with a distance loss function}. As a result, the connected edges in SRG are assigned by the pairwise distances between feature representations.

Instead of naively traversing the graph and finding the shortest path, which potentially prevents us from exploring a new sequence, we find paths by the proposed SDPF algorithm. Conceptually, SDPF firstly estimates the candidates, which are potential to construct a new sequence, and then distills them through constraints to define the final node at each path-finding-iteration. Finally, the sequence of nodes in the path is mapped to the corresponding frames to produce smooth video results. We subsequently elaborate on each module.

\section{Graph generation with RSFNet}
Given a set of frames, we now aim to build a complete graph of this set prior to the resequencing manner. \textcolor{black}{As mentioned in the related work, we propose a network RSFNet to get knowledge on their feature representation and embed the samples to a specific metric space where the similarity or distance between any two samples is clearly represented. Once the distance metric is learned, feature representations and distance are capable of reflecting the relation of input images. RSFNet is a reusable structure that reduces the computational cost and efficiently represents image information. RSFNet shoulders two significant roles in the graph generation: firstly, RSFNet calculates latent vectors corresponding to the given frames. Each vector is treated as a node in the graph. Secondly, RSFNet is trained with a proposed distance loss to infer the similarity of latent vectors. This manner facilitates the distance of latent vectors more accurate. The details of the proposed framework are presented as follows.}

\input{Encoder}
\subsection{RSFNet Structure} 
In a common Convolutional Neural Network (CNN) framework, an encoder converts the \textcolor{black}{input image $x$ into a representation vector $r$ ($r = \Phi (x)$)}. The architecture of an encoder $\Phi(.)$ depends on the input in a specific application. For instance, in the application of image classification, the CNN is a good choice. When applied to video resequencing, such an off-the-shelf CNN might not be suitable since contextual information in a specific frame is necessary for generating new sequences. \textcolor{black}{Besides, human often }relies on a high-level semantic understanding of the video contents, usually after viewing the whole sequence, she/he can decide which frame should be selected the next frame in the sequence. Therefore, it is necessary to differentiate the target sequence scene to make the resultant sequence semantic, reasonable, and smooth. At this point, an encoder with a pure CNN structure may lack sufficient information for such an appealing sequence.

Motivated by the above reason, we design our RSFNet \textcolor{black}{by the triplet of \textbf{R-Encoders}, which share the parameters, i.e., weights and biases. Each R-Encoder consists of two modules, \textit{Coarse Feature Extractor} (CFE) and \textit{Recursive Feature Fining} (RFF). The design is visualized in Fig.\ref{fig_encoder}}. For the CFE module, we treat it as an extractor to obtain the initial feature maps. The backbone network of CFE is based on the VGG-19 network \citep{16_simonyan2014very}. This pre-trained network is widely used in several applications in a feature extraction manner. Hence, it is reliable to be considered a good feature extractor. Furthermore, VGG-19 has been trained on a large-scale dataset. With this strategy, we can reduce the burden in training for this process. We solely use the first four blocks and remove the fifth block from the original VGG-19 framework since it lacks pixel-wise content information  \citep{18_gatys2016image}. An input image $\mathcal{I}$ is firstly fed to CFE. Let matrix $\mathcal{X}_i \in \mathrm{R}^{H_i \times W_i \times k}$ denote corresponding feature maps produced by four layers of CFE. Here, $k$ is the number of channels of each feature response. $H_i$ and $W_i$ are respectively the height and width of the feature maps in layer $i$ ($i = 1 \dots 4$). As shown in Fig.\ref{fig_encoder}-(b), the feature maps $\mathcal{X}_i$ are enhanced along the channel and space dimensions to obtain the feature maps $\mathcal{F}_i$ by RFF module. In other words, instead of directly utilizing feature maps from CFE, we propose an RFF module to integrate with CFE to \textcolor{black}{produce features that can depict the variety content in frames}. The effectiveness of this design is visualized by \textcolor{black}{the analysis} in the later session \textbf{A.1}. 

The \textbf{\textit{Recursive Feature Fining}} (RFF) module is the core of \textcolor{black}{an R-Encoder}, which shoulders the task of preserving \textcolor{black}{contextual information} of images during encoding into latent space. RFF is formulated by recursively integrating feature maps of CFE. \textcolor{black}{A straightforward technique could be used instead of RFF is that re-scaling the feature maps obtained from CFE and combining them together. However, the feature extraction from a backbone, \textit{e.g.} either VGG or ResNet, is performed by a repeated process of convolutional and max-pooling operations. These extracted features by themselves loss the low-level information that is likely to aid in discriminating object regions from the background regions. Thus, such a simple technique, \textit{i.e.}, re-scaling, might neglect smaller objects or information in the background regions and eventually decrease the capability of the encoder. In the structure of our RFF, we} embed two blocks, NCM and C1R. NCM is to normalize the input feature maps before the concatenation. Meanwhile, the C1R block's task is to compress the size of feature maps without losing information.


The network architecture of the proposed RFF is shown in Fig.\ref{fig_encoder}-(b). Four feature maps with different resolutions obtained from CFE ($\mathcal{X}_i$) are taken as the inputs of the RFF. Mathematically, the above process can be recursively expressed as:
\begin{equation}
    \begin{cases}
      \mathcal{F}_i = \Psi(\varphi(\mathcal{F}_{i-1} \otimes \mathcal{X}_i)), (i = 2 \dots 4) \\
      \mathcal{F}_1 = \Psi(\mathcal{X}_1)
    \end{cases},       
\end{equation}where $\Psi(.)$ and $\varphi(.)$ denote the functions from the NCM block and C1R block, respectively; $\otimes$ is the concatenation operation. By concatenating two different feature maps, resultant feature maps $\mathcal{F}_i$ $(i>1)$ simultaneously captures two different receptive fields.

To be more specific, NCM is designed to enhance the spatial representation for the input feature maps from VGG-19. This block performs the \textit{Normalization$\rightarrow$Conv3$\times$3$\rightarrow$MaxPooling} structure.
Output of input feature maps $\mathcal{F}^{in}$ passed through NCM is performed as
\begin{equation}
    \Psi(\mathcal{F}^{in}) = P(C^3(Norm(\mathcal{F}^{in}))),
\end{equation} where $P(.)$ represents the Max-Pooling operator; $C^3(.)$ indicates the standard convolution with the kernel size of $3\times 3$; and $Norm$ is a normalization operator.

C1R employs a $1\times1$ point-wise convolution and a residual block. Our residual block consists of two batch normalization (BN) layers and two $3\times3$ convolutional layers. Note that, compared with the basic residual block \citep{17_he2016identity}, our residual block removes the RELU layer after the first convolutional layer to preserve more spatial details. See Fig.\ref{fig_encoder}-(b), immediately after the concatenation which is used to transmit the information of these two distinct layers, this block is embedded to learn the correlation of feature maps from different layers. This process is expressed as:
\begin{equation}
    \varphi(\mathcal{F}_c) = C^1(\mathcal{F}_c) + BN(C^3(BN(C^1(\mathcal{F}_c)))),
\end{equation}where $\mathcal{F}_c$ is the resultant feature maps after the concatenation phase. $C^1$ represents the $1\times1$ point-wise convolution. 

With our above design, some benefits can be gained. First, using a pre-trained network as a backbone significantly reduces the training cost. Second, RFF can be easily embedded into an existing neural network. In the design of C1R block, $1\times1$ convolution is to increase channels corresponding to the previous layer. Meanwhile, residual connection sufficiently mitigates the gradient vanishing problem, which usually occurs when training the deep network. 

\textcolor{black}{We need to build the embeddings of frames \textcolor{black}{such that} they have the following properties: (1) two similar frames produce two embeddings so that the mathematical distance between them is small, and (2) two very different frames produce two embeddings so that the mathematical distance between them is large. To do that, we model RSFNet that contains the triplet of R-Encoders, which use the same weights while working in tandem triplet of different input vectors to compute comparable output vector. In our training, the distance loss $\mathcal{L}_d$ is used as the objective function to reinforce the distance between two latent vectors to match the similarity of the images well reflect pixel-level image similarity. We train RSFNet using a set of triplet images - an anchor $x_a$, its positive $x_p$, and negative $x_n$. Detail of preparing such triplet data is discussed in our supplementary material. For three embeddings $r_a, r_p, r_n$ of the images $x_a, x_p, x_n$, respectively, the formula of the distance loss is as follows. 
\begin{equation}
    \mathcal{L}_d = -z \log \big(\xi\big) + (z-1) \log \big(1-\xi\big), 
\end{equation}
where 
\begin{equation}
\xi = \gamma(\parallel r_a - r_n \parallel_2 - \parallel r_a - r_p \parallel_2), 
\end{equation}
and
\begin{equation}
z=\begin{cases}
			1, & \text{if $d_p(x_a, x_n) > d_p(x_a, x_p)$}\\
            0, & \text{if $d_p(x_a, x_n) < d_p(x_a, x_p)$}
		 \end{cases}.    
\end{equation}
Here, $\gamma(.)$ is the sigmoid function \citep{nwankpa2018activation}, used here to ease the severe gradient problem. $d_p(.)$ is the PSNR measurement \citep{hore2010image}; Eq.(6) is used to get the initial knowledge about the similarity of image pairs. It indicates which term in Eq.(4) will be visible during the training. Therefore, $z$ can be treated as an indicator to determine whether $x_a$ is similar to $x_n$ or $x_p$. We note here that using any pixel-level distance metric in Eq.(6) could yield an equivalent effect. It is clear that Eq.(4) encourages the embedding of $x_a$ to be closer to $x_p$ than to $x_n$. \textcolor{black}{Optimizing these terms boosts the margin between distances of negative pairs and distances of positive pairs.} The effectiveness of this formulation is discussed by ablated results in later session \textbf{A.2}.}

\subsection{Learning-Based Euclidean metric}
To define the weight of each edge in the complete graph, we calculate the distance of all pairs of latent vectors. The distance metric used in this manner should satisfy two criteria: (1) it can \textcolor{black}{well} reflect the distance of images, i.e., a distance in low dimensional space should be consistent with the \textcolor{black}{content correlation of images}, and (2) not too expensive to reduce the burden when the number of given images is significant. Our early experiments considered five different distance metrics: the Hausdorff distance, Earth Movement Distance (EMD), the LPIPS distance, SSIM, and the Euclidean distance. However, Hausdorff and EMD have a good performance on specific data, i.e., cartoon characters \citep{de2004cartoon, 3_yang2010recognizing}. LPIPS is an expensive computation metric, it takes approximately five seconds on an image pair. \textcolor{black}{SSIM and Euclidean distance metrics are potential. However, Euclidean metric is the most common use of distance measure and known as simple distance. When data is dense or continuous, this is the best proximity measure. Thus, we consider Euclidean as the baseline in learning the relation of images in our current application. It's worth noting that directly using pixel-level distance metrics, such as SSIM or Euclidean, without Eq.(4) is not sufficient in our current application. The reason is that we target to explore new transitions on the diverse content frames. Simply employing a plain distance metric without the objective function $\mathcal{L}_d$ prevents us to reach this goal. This could be seen in the ablated visualization \textbf{A.2}}

Instead of directly using Euclidean distance to measure the metric value between two features, in RSFNet, we apply deep learning technique to further learn their similarity. When the self-defined metric space is an Euclidean space, the metric value between two samples is a distance metric, which is defined as:\textcolor{black}{
\begin{equation}
    d_{ij}(v_i, v_j) = \parallel \mathcal{R}(x_i) - \mathcal{R}(x_j) \parallel,
\end{equation} where $\mathcal{R}$ is our trained RSFNet; $x_i$ and $x_j$ are the corresponding frame of embedding $v_i$ and $v_j$, respectively.}

\section{Single-source Distillation path-finding}
In this section, we present our approach of finding the path on the complete graph to construct new sequences. Let $\Omega$ be the set of latent vectors $v_i$ obtained from our RSFNet and $d_{ij}$ be the distance between two latent vectors $v_i, v_j \in \Omega$ \textcolor{black}{defined by Eq. (7)}. We construct a graph $\mathcal{G} = (\mathcal{V}, \mathcal{E})$ in which each node $V_i \in \mathcal{V}$ represents a latent vector $v_i \in \Omega$ and the weight of each direct edge $e_{ij} \in \mathcal{E}$ (from $V_i$ to $V_j$) is assigned by the corresponding distance $d_{ij}(v_i, v_j)$. 

Once graph $\mathcal{G}$ is constructed, our system lets the user choose a node randomly. An expected sequence can be constructed by traversing the graph starting from this node. A possible and straightforward way is finding the shortest path on the graph with the selected node because the edge of a pairwise node reflects their similarity, i.e., if the weight of an edge is smaller, the connected nodes are more similar and vice versa. Hence, this naive strategy is tolerant of plausible sequences if the input clips do not have dense motion and content.

The question here is - \textit{How do we construct the sequences  \textcolor{black}{that are different from those in the input video}?} Resequencing videos without pre-processing (e.g., extracting objects from the background), we may face a range of challenges in image content (e.g., the video has multiple moving objects, dense motion directions, or with complicated background). \textcolor{black}{Generating new sequences} while avoiding flicking artifacts, such a classic shortest path-finding technique by itself is not tailored. The reason is that the resultant path found by this technique, such as \citep{morace2022learning}, tightly reflects the similarity of the sequence in the given clip. It may yield a similar sequence to the given sequence. Otherwise, it may fall into chaotic motion if the clip has dense movements. \citet{3_yang2010recognizing} tackle this issue by extracting the cartoon character from the image content and using the motion direction feature (MDF) to evaluate the gesture dissimilarity. However, they focus on the frames that have a single cartoon character. If the frames demonstrate the motion of multiple objects, using MDF might be insufficient. Recent work by \citet{morace2022learning} also suffers from this issue if there exist dense motion directions (Chinese ink in Fig.\ref{fig_testing_set}-(K)). To overcome these challenges and \textcolor{black}{produce new sequences}, we propose an algorithm called \textit{Single-source Distillation Path-Finding} (SDPF) to find the path when traversing on the graph.

When designing SDPF, we base on the fact that the adjacent frames have to be consistent in content information and temporal coherency in a particular clip. Thus, in finding a \textcolor{black}{new }path that satisfies these two aspects, we consider them whenever choosing a node at every step. We call the phenomenon, which is caused by missing one of the two aspects, as a \textit{cold transition}. More specifically, we model our SDPF to work under the control of two-layer distillation. Given a graph and a starting node, the first layer is to distill the set of candidates, which are the potential to be consistent with the content. Taking this set as input, the second layer estimates the plausible motion direction that could be generated and distill the candidates on the set that are potentially temporally coherent. In the following, we call the current node $V_c$; our SDPF aims to find the adjacent node of $V_c$, denoted as $V_{c+1}$. We \textcolor{black}{visualize} the difference of Single-source shortest PathFinding (SSPF) versus our proposed SDPF in Fig.\ref{fig_LSM_visualize}(a), (b). SSPF chooses \textcolor{black}{only one} node, which has the shortest cost, to add to the path. In contrast, our SDPF considers \textcolor{black}{a number of nodes, \textit{e.g., }three nodes in this example}, which have the cost lower than a designated threshold, as the potential candidates in equivalent probability to be added to the path.

There are several benefits of using our SDPF algorithm. First, we can explore new paths since we do not strictly follow the theory of the shortest path. Second, we can control the motion direction to be locally consistent in clip segments and globally realistic in the generated clip. Third, it is faster than such a greedy strategy. We subsequently describe how we model the constraints in our SDPF algorithm. The pseudo-code of SDPF is presented in Algorithm \ref{SDPF}.

\begin{algorithm}
\caption{SDPF Algorithm}
	\begin{algorithmic}[1]
	\Statex \textbf{Input: } Set of latent vectors $\{v_i\}$, distance metric $\{d_{ij}\}$
	\State $\mathcal{V} \leftarrow \{v_i\}$, $\mathcal{E} \leftarrow \{d_{ij}\}$;
	\State Construct graph $\mathbf{G} = (\mathcal{V}, \mathcal{E})$;
	\State $V_o \leftarrow $ user's selection; 
	\State Initialize a list $\mathcal{P}$ to subsequently push the selected node to the path;
	\State Add $V_o$ to $\mathcal{P}$
	\State $V_c \leftarrow V_o$; \textcolor{gray}{\textit{\small/* $V_c$ is the node at current state*/}}
	\Statex \textcolor{gray}{\textit{\small/* Distillation in the first layer*/}}
	\For{each node $V_j \in \mathbf{G}(\mathcal{V}-\mathcal{P})$}
	    \If{$e_{cj} < \eta$}  \textcolor{gray}{\textit{\small /* $\eta$ is defined in Eq.(8)*/}}
	        \State Add $V_j$ to $\mathcal{S}_1$   
	    \EndIf
	\EndFor
    \Statex \textcolor{gray}{\textit{\small /* Distillation in the second layer */}}
    \For{each node $V_k \in \mathcal{S}_1$}
	    \If{$V_c \in LMS $}
	        \State $\mathcal{S}_2 = C_d(V_c, V_k) + C_t(V_c, V_k)$
	   \Else 
	        \State $\mathcal{S}_2 = C_t(V_c, V_k)$
	    \EndIf
    \EndFor
    \For{each $V_i \in \mathcal{S}_2$} 
        \State Compute possibility $\Omega$ for each $V_i$ by Eq.(20);
    \EndFor
    \State Choose $V_i$ by randomly selecting $\Omega$;
    \State Add $V_i$ to path $\mathcal{P}$;
    \State Update $V_c \leftarrow V_i$
    \Statex \textbf{Output: } Sequence of path $\mathcal{P}$
	\end{algorithmic} 
\label{SDPF}
\end{algorithm}

\subsection{Content-aware distillation}
In this layer, we find the set of candidates that have relevant content to the current node $V_c$ rather than finding the node that have smallest distance to $V_c$. Obviously, if $V_{c+1}$ is the node that has the smallest weight to $V_c$ among the directed nodes of $V_c$, this may yield the resultant sequences \textcolor{black}{that are similar to the source sequences}. Thus, we find the candidates that are potential to explore new \textcolor{black}{transitions}. This saves the generated video from flicking artifacts due to the \say{\textit{cold transition}} between them. We construct a set $\mathcal{S}_1$ of candidates that are relevant to $V_c$ as:
\begin{equation}
    \mathcal{S}_1 = \{V_i \in \mathcal{G}: e_{ci} < \eta; \text{s.t. }\eta = \frac{\sum e_{ij}}{N}\},
\end{equation} where $N$ is the total number of nodes in the graph $\mathcal{G}$. In this equation, $\eta$ is the threshold that represents for the mean of the weights in the graph $\mathcal{G}$. By this configuration, an edge has the weight that smaller than $\eta$ could be considered as a \say{\textit{potential candidate}}. \textcolor{black}{It is hypothesized that we set another variable such as top $k\%$ of the candidates that have the closest weight to the minimum weight of the graph, the size of $\mathcal{S}_1$ is increasing with the total number of frames in the given video. If the clip is short, the size of $\mathcal{S}_1$ is small, and thus it might be not sufficient to explore a new path. If the clip is long, the size of $\mathcal{S}_1$ accordingly increases, and thus it might include the wrong candidates (i.e., the candidates are not correlated). Therefore, the threshold $\eta < mean(.)$ is tolerant with different amounts of frames and able to avoid these phenomena.}

With Eq.(8), we can eliminate the nodes that have low reliability and drive our focus on the nodes that are highly potential to be content correlation. Each element of $\mathcal{S}_1$ represents for a possible way that we can explore from $V_c$ without suffering from \textit{cold transition}. Note that the size of $\mathcal{S}_1$ varies along with $V_c$ at each iteration. And in $\mathcal{S}_1$, the nodes are treated equivalently, i.e., the probability of choosing a node is independent with the edge weight.

\subsection{Motion direction-aware distillation}
Having computed the set $\mathcal{S}_1$, the next question is - \textit{which candidates in this set can yield a plausible motion direction?} Note that here \say{\textit{plausible}} refers to both backward motions, forward motions, or any movement but avoiding the results from the flip-flop and jumping phenomena. To do this, we propose to our SDPF algorithm two constraints: \textit{Directional distillation} and \textit{Coherent distillation}. Detail of each constraint is described as follows. 

\subsubsection{Directional distillation}
This constraint, denoted as $C_d$, is proposed to control the consistency of motion direction. To achieve this, linear motion segment (LMS) is the factor we consider here. LMS is ubiquitous in real-world videos. Readers can see the visual example of LMS here \footnote{\href{http://graphics.csie.ncku.edu.tw/SDPF/LMS.mp4}{http://graphics.csie.ncku.edu.tw/SDPF/LMS.mp4}}. \textcolor{black}{Resequencing such videos may fall into two kinds of motion-noise: (1) flip-flop phenomenon due to the LMS-frame is not recognized, causing inconsistent direction, and (2) abnormal motion since both backward and forward motions yield smooth transitions. Therefore, there is a need to recognize the major motion direction in frames as well as detect the LMS to avoid these motion-noises.}

\input{LMS_visualize}

Let $\mathcal{X} = \{x_a\}^n_{a=0}$ be the sequence of frames $x_a$ in the given video, $n$ is the number of frames, we first calculate the optical flow \citep{21_sun2018pwc} of $\mathcal{X}$ and denote this set as $\mathbf{Y}=\{F_{a\rightarrow a+1}\}^{n-1} _{a=0}$, \textcolor{black}{here $F_{a\rightarrow a+1}$ is the optical flow of frame $x_a$ to $x_{a+1}$}. To focus on drastic changes in the optical flow, we normalize each element in $\mathbf{Y}$ as follows:\textcolor{black}{
\begin{equation}
    \widehat{\mathbf{N}_{ij}} = \frac{\parallel F_{ij} \parallel_2 - \min_{ij}\parallel F_{ij} \parallel_2}{\max_{ij}\parallel F_{ij} \parallel_2 - \min_{ij}\parallel F_{ij} \parallel_2}.
\end{equation}}\textcolor{black}{We denote this set as $\widehat{{\mathbf{Y}}} = \{\widehat{\mathbf{N}^a}\}^{n-1} _{a=0}$. Video frames may have various motions, \textit{e.g.}, motions of main object(s) or light motions of background objects. To recognize the major motion direction in frames, we mask on each frame a value called \textit{\say{motion tendency}} ($T$), as shown in Fig.\ref{fig_LSM_visualize}(c). This value represents for the motion direction that dominates in a frame, which is formulated by average normalized vectors of partial optical flow:
\begin{equation}
    T = \angle \Bigg(\frac{\sum{\widehat{\mathbf{N}_{ij}}}}{n \times m} \Bigg), \text{ \textit{s.t.,} } \widehat{\mathbf{N}_{ij}}> \sigma \frac{F_{ij}}{\parallel F_{ij} \parallel_2},
\end{equation}}where $m, n$ is the width and height of frames, respectively. Threshold $\sigma$ is set to 0.5 in our experiments to ensure only huge changes to be concentrated.

\textcolor{black}{Thereafter, we base on motion tendency in frames to detect LMS-frame, the frame that belongs to an LMS. The definition of an LMS-frame is expressed as:}
\begin{equation}
    \exists j, k \in \mathrm{N}: j \leq i \leq j+k, k>2, \text{ s.t. } |T_j - T_l| \leq \delta,
\end{equation}for all $l\in [j, j+k]$. In our experiments, we compute motion tendency of frames and mask them with motion tendency value if the frames belong to LMS prior of path-finding manner, and threshold $\delta$ is set to $\frac{\pi}{4}$.

\textcolor{black}{Finally, we configure the constraint for directional distillation as:
\begin{equation}
    C_d = \big |T_c - T_k \big | \leq \xi, \text{ \textit{if} } x_c \in \text{LMS} \& \exists V_k \in \mathcal{S}_1  \text{ \textit{s.t.,} } x_k \in \text{LMS},
\end{equation} where $T_c, T_k$ is the motion tendency of the corresponding frame $x_c, x_k$ of the node $V_c, V_k$, respectively, $k \in [1 \dots n_1]$, and $\xi$ is set to $\frac{\pi}{3}$. Here, $n_1$ is the size of set $\mathcal{S}_1$. The condition in Eq.(12) reveals that constraint $C_d$ only works if the corresponding frame of $V_c$ is an LMS-frame \textcolor{black}{and there exist an LMS-frame in $\mathcal{S}_1$}. Otherwise, we skip this constraint. \textcolor{black}{The visual sample can be found in Fig.\ref{fig_LSM_visualize}(d). We can see that, $V_c$ and two of its three candidates are LMS-frames. In this case, Eq.(12) is used to avoid flip-flop phenomenon.}} We analyze the effectiveness of this constraint with ablated results in session \textbf{A.3.1}.

\subsubsection{Coherent distillation}
The distillation in this layer is proposed to maintain the temporal coherency in generated sequences. \textcolor{black}{\citet{3_yang2010recognizing} extract cartoon characters from frames and compute the angle of two motion direction features of the characters to define the differences of motions. In the cases that video frames consists of multiple moving objects, this technique is not practical. We instead propose a \textit{Pixel-wised Motion Similarity Measurement} (PMSM) to shoulder the smoothness of generated sequences. }

\textcolor{black}{As named, PMSM measures the pixel-wise motion similarity between two frames. To get knowledge of motion in frames, inspired by \citep{jiao2021optical}, we use optical flow as the motion feature. Thus, a possible and straightforward way we can measure the motion differences is using optical flow directly. Nonetheless, as aforementioned, various motions of multiple objects in frames cause challenging to define the consistency between them. We therefore learn the motion feature by mapping optical flow domain to image domain. In other words, given two frames, we use the corresponding the optical flow of these frames to construct the instance in image domain, dubbed \textit{pseudo-image}. A pseudo-image is made by the major motions in the corresponding frame and the correlated motion of frame-pair. We finally calculate the distance of pseudo-images to measure how smooth the motion changes in a transition. A smaller PMSM reveals a smooth transition. Consequently, we use PMSM to configure the constraint in this distillation layer, so-called $\mathbf{C}_t$, to control the motion in adjacent frames not to change frequently or drastically.}

\textcolor{black}{Fig.\ref{fig_DOF} outlines the flowchart of PMSM. For each node $V_k$ in the set $\mathcal{S}_1$, we treat it as a hypothesized adjacent node of $V_c$. And $x_c$, $x_k$ respectively are the corresponding frames of node $V_c$, $V_k$. The smoothness of transition from frame $x_c$ to $x_k$ is now defined by the motion distance of two optical flows FC and FK, where FC is the optical flow of frame $x_c$ to its backward adjacent frame in the input video. The reason for the order of this calculation is explained in detail in our supplementary. Similarly, FK is the one of $x_k$.}

\textcolor{black}{With two optical flows FC and FK, we first normalize their magnitude by Eq.(9), denoted as $\widehat{\text{FC}}$, $\widehat{\text{FK}}$, respectively. We then define a map of significant motions with:
\begin{equation}
    \mathcal{M}_{ij} = \max(\widehat{\text{FC}}_{ij}, \widehat{\text{FK}}_{ij}),
\end{equation}with $i=0 \dots W$, $j = 0 \dots H$; $W, H$ is the width and height of the frame, respectively. The map $\mathcal{M}$ represents for the correlation of major motions in frame-pair. We get these information to learn how to control the pixel-wise consistency in the pseudo-images. }
\input{PMSM}

\textcolor{black}{In each normalized optical flow $\widehat{\text{FC}}$, $\widehat{\text{FK}}$, we count the number of elements that are larger than a threshold $\mu$. We denote as $E = \{e_1, \dots, e_{N_e}\}$. Note that the size of set $E$ varies along $\widehat{\text{FC}}$ or $\widehat{\text{FK}}$.} Since we only use $N_e$ optical flows in $\widehat{F}$ to model $C_t$, value of $N_e$ should be large enough. A small $N_e$ will decrease the difference between optical flows. This yields to that we may fail to define the difference correctly. Therefore, if $N_e$ is smaller than a threshold, i.e., 224 is the height and width of frames, we will cut $\mu$ in a half and compute $\mu$ again to ensure $N_e$ is sufficient. Initially, we set $\mu$ to $\frac{1}{2}$. 

Thereafter, we rely on $\mathcal{M}$, $E$ to map back to the input optical flow to construct pseudo-image. More specifically, $\forall i,j$ in optical flow $F_{ij} = (x_{ij}, y_{ij})$, pseudo-image $\mathbf{X}^p \in \mathbb{R}^{H\times W \times 3}$ is expressed as:
\begin{equation}
    \mathbf{X}^p_{ij} = \begin{cases}
    \big(\frac{x_{ij}}{2 \parallel F_{ij} \parallel_2} + \frac{1}{2},  \frac{y_{ij}}{2 \parallel F_{ij} \parallel_2} + \frac{1}{2}, 1\big), \text{ if } \mathcal{M} \geq e_k \\
    \big(\frac{1}{2}, \frac{1}{2}, 0 \big), \text{ if } \mathcal{M} < e_k
    \end{cases},
\end{equation}where $e_k$ is the largest element of $E$ in $\mathcal{M}$. \textcolor{black}{In Eq.(14), if the parameters are $\mathcal{M}$, $E_c$, and $\widehat{\text{FC}}$, we can construct pseudo-image of node $V_c$, denoted as $\mathbf{X}^p_c$. Similarly, we can get the pseudo-image $\mathbf{X}^p_k$ from those of node $V_k$. The first two channels in $\mathbf{X}^p$ are the unit vector of $F$ with constant translation, in which unit vector provides only direction information. The constant translation makes the value to be in range of $[0, 1]$ without any computation error. The third channel is used to enlarge the difference between the feature point and other pixels. }

At the end, motion distance of two optical flows is formulated as the similarity of the corresponding pseudo-images:
\begin{equation}
    \delta(FC, FK)= 
      -\parallel \mathcal{R}(\mathbf{X}^p_c) - \mathcal{R}(\mathbf{X}^p_k) \parallel_2,
\end{equation}where $\mathcal{R}(.)$ indicates our trained RSFNet. \textcolor{black}{In essence, pseudo-images have different appearance compared to video frames, \textit{i.e.}, pixel value represents for the motion intensity of objects in the corresponding frame. Encoding such pseudo-image serves the knowledge of the regions that have considerable motions}. It's worth noting that motion distance of two frames in Eq.(15) also could be expressed by the similarity of pseudo-images. However, to make $\delta(.)$ stable when working on diverse motions, we feed them to RSFNet. Although RSFNet is trained on video frames data, RSFNet on the other hand learn a similarity function to see if two images are the same. This enables to discriminate new classes of data without training the network again. We give out discussion and visualization on these effects in the supplementary file.

Equation (15) represents the relation adjacent frames in term of motion change degree. For each node $V_k$ in the set $\mathcal{S}_1$, $k = 0, \dots, n_1$, we define constraint $C_t$ as:
\begin{equation}
    \delta(FC, FK) \leq \omega,
\end{equation} where $\omega$ is set by:
\begin{equation}
    \omega = \begin{cases}
      \frac{1}{n_1}\sum_{k \in \mathcal{S}_1}\delta(FC, FK), \text{ if }n_1 \geq 2 \\
      \min\big(\delta(FC, A_1), \delta(FC, A_2) \big), \text{ if }n_1 < 2
    \end{cases},
\end{equation}here $A_1$ and $A_2$ are the augment form of FC, i.e., $A_1$ is the rotation of FC with angle $\frac{1}{2}\pi$ and $A_2$ is the rotation of FC with angle $-\frac{1}{2}\pi$. We set $\omega$ as the average difference of $\mathcal{S}_1$ is intuitive. However, the average will loss its function if the number elements in $\mathcal{S}_1$ is less than 2. Therefore, we calculate the difference between FC and the rotation of itself to ensure the direction of the motion is sufficiently smooth. We analyze the effectiveness of this constraint by the ablated results in later session \textbf{A.3.2}.

In summary, the constraint model of distillation in this layer can be factorized as:
\begin{equation}
    \mathbf{C}_d\big(V_c, V_k\big) + \mathbf{C}_t\big(V_c, V_k\big),
\end{equation}where $V_k \in \mathcal{S}_1$. In the cases that $V_c$ does not belong to LMS or there does not exist a candidate in $\mathcal{S}_1$ that belongs to LMS, the first factor in this equation is omitted. In other words, we define the candidates that we can add to the path as:
\begin{equation}
    \mathcal{S}_2 = \begin{cases}
      \mathbf{C}_d\big(V_c, V_k\big) + \mathbf{C}_t\big(V_c, V_k\big), \text{ if } V_c \in LMS \\
     \mathbf{C}_t\big(V_c, V_k\big), \text{otherwise}
    \end{cases}
\end{equation}

\subsubsection{Final selection}
Thus far, the candidates in $\mathcal{S}_2$ are \textcolor{black}{the possible nodes we can choose to explore. In the cases that $\mathcal{S}_2$ is empty, the algorithm will early stop to maintain the quality of resultant clips. If $\mathcal{S}_2 > 1$, we adopt Softmax parameterization protocol \citep{goodfellow2016deep} to converge the selection in each iteration. Let $\delta_j$ be the motion distance from a node $V_j \in \mathcal{S}_2$ to $V_c$, we parameterize the possibility of choosing $V_j$ as:
\begin{equation}
\Omega(V_j|\mathcal{S}_2) = \frac{\text{exp}(\delta_j)}{\sum^{n_2}_{i=1}\text{exp}(\delta_i)},
\end{equation} where $n_2$ denotes the number of candidates in $\mathcal{S}_2$, \textcolor{black}{$\delta_i$ is the motion distance of node $V_i$ and $V_c$}. This equation is used to compute the possibility of a vertex to be chosen. Then, we select the adjacent node of $V_c$ according to randomly choose the possibility $\Omega$. It's worth pointing that choosing any candidates in $\mathcal{S}_2$ is sufficient to guarantee smooth and plausible sequence. However, we aim to explore the novel path, we thus utilize Eq.(20) to increase the possibility of sequencing novelty rather than choosing the smallest edge-weight node. Furthermore, this strategy enables users to have more predictable results. The efficiency of this design is visualized by video results in the supplementary video.}

\section{Experimental Results}
\subsection{Implementation Details}
We implemented our proposed resequencing system in Tensorflow \citep{abadi2016tensorflow}. All experiments were performed on a PC equipped with Intel Core i7-770 CPU, 16GB RAM and an NVIDIA GTX 1070 GPU. The User Interface (UI) is developed by QT toolkit \citep{eng1996qt}. We train our model with patch size of 8. Adam optimizer \citep{20_kingma2014adam} is used. Early-stopping with 10 epochs patience is used to prevent over-fitting. To reach the minimum of loss, we cut the learning rate in half when the validation loss does not improve in 3 epochs.

\subsection{\textcolor{black}{Our results and discussion}} 
\textcolor{black}{Fig.\ref{fig_testing_set} exhibits the frames of some typical videos in our experiments. Readers are encouraged to explore our project website\footnote{\href{http://graphics.csie.ncku.edu.tw/SDPF/}{http://graphics.csie.ncku.edu.tw/SDPF}} to access more visual results. The aspects that make our results and system advance prior works could be summarized as follows. }

\textcolor{black}{We are capable of resequencing both cartoons (Fig.\ref{fig_testing_set}-(A) to (G)) and natural videos (Fig.\ref{fig_testing_set}-(H) to (L)). Cartoon images often consist of sharp lines, flat backgrounds, and smooth color blocks, while natural images contain more complex and local textures \citep{chen2019blind}. This ability is adopted by benefiting of the proposed RSFNet and the distance loss. RSFNet boosts the performance of our system in understanding high-level features of natural images; meanwhile, the distance loss facilitates the accuracy of image feature-pairs similarity.}

\textcolor{black}{We are capable of resequencing the clips, which consist of \textcolor{black}{complex motions, i.e., the motion of multiple objects or dense motion directions.} This aspect is adopted by the \textit{Motion Direction-Aware Distillation} in our SDPF algorithm. As examples, let us
take Fig.\ref{fig_testing_set}-(F) and Fig.\ref{fig_testing_set}-(G). The challenge here is that both cases consists of multiple simultaneous motions.} In Fig.\ref{fig_testing_set}-(G), we have to \textcolor{black}{control} the consistency of movements of two objects: 1) the direction when the bear raises his hand to hold the flower and rotates it, and 2) the other flower waves with the wind. Meanwhile, in Fig.\ref{fig_testing_set}-(F), such a resequenced clip should maintain the consistency of the movements of the lady, baby, the car, and the windshield wipers. Nevertheless, we can generate appealing results, \textit{i.e.}, we re-sequence the new clips without damaging the coherency and flicking artifacts. The challenge also falls in the natural scenes here. These samples encompass linear motion segments, which cause resultant sequences to be a flip-flop phenomenon. Thanks to the constraints embedded in our SDPF, we revolve this challenge and produce smooth transitions.

Another interesting aspect of our system is the ability to produce the sequences which \textcolor{black}{are different} to those in the given clip. This aspect is adopted by the \textit{Content-Aware Distillation} in our SDPF Algorithm. More specifically, we visualize the filmstrips of two paths which are from the original video and our result in Fig.\ref{fig_different_sequences}. We can see with the same image gallery, but our sequence is quite different from those in the input video. By observing this resequencing result, we can see the transition of each single frame pair is plausible. The full clips can be seen in our supplementary videos.

In addition, we can generate different sequences according to the starting frame, which the user selects. This aspect enables users to obtain various predictable results. \textcolor{black}{Fig.\ref{fig_different_sequences} is a sample. More results can be seen on our project website. By observing the filmstrips in the figure, the sequence generated by our method is not only relatively different from the source sequence but also smooth in transitions.}

\input{Testing_set}
\input{different_sequences}
\input{Differences_Frames}

\subsection{Evaluation Metrics}
\textcolor{black}{To evaluate the performance of the proposed method, we measure the generated sequences with three aspects: (1) the stability of videos, (2) the difference degree of generated sequences, and \textcolor{black}{(3) human perception on our results}. In this evaluation manner, we totally use 12 videos (shown in Fig.\ref{fig_testing_set}), which are rendered from our system. Then, we synthesize them for the below evaluation metrics.} 

\subsubsection{\textbf{E.1.} Stability measurement}
As the generated videos are explored according to the user's selection of the starting frame, they may not have the ground truth. To measure the stability of rendered videos, we synthesize 12 videos by our method and the corresponding source video; and measure the differences between adjacent frames. The reason is that the source videos by themselves are \textcolor{black}{temporally} coherent; our results are rendered from the same image set with them \textcolor{black}{but probably in different orders}. Thus, we treat them as the standard to \textcolor{black}{judge the stability degree} of the results. 

\input{Charts}
Given two adjacent frames $F_{t-1}$ and $F_{t}$, difference of them is factorized as:
\begin{equation} 
    \mathcal{D}_{t \rightarrow t-1} = \parallel F_t - F_{t-1} \parallel, 
\end{equation}here $t \in [0 \dots N_s]$, $N_s$ is the total frames in the video. After that, we calculate the mean ($M_{\mathcal{D}}$) of $\mathcal{D}_{t \rightarrow t-1}$, and we compare them against those in the source video. \textcolor{black}{On each single pair, $D_{t \rightarrow t-1}$ in our result might be higher than those in the ground truth, but it should be at an acceptable rate to guarantee there does not exist notable flicking artifact. This eventually affects the quality of the entire rendered clip. Therefore, we base on $M_D$ to judge the stable quality of the results, i.e., the more tightly asymptotic to those in the source is better.} \textcolor{black}{We visualize an example of this manner in Fig.\ref{fig_differences_frames}.} In this visualization, we choose a mutual frame between our rendered clip and the corresponding source video (i.e., frame 9). \textcolor{black}{We can observe that the adjacent frames of frame 9 in the source (i.e., frame 11) and the adjacent candidates (i.e., frames 16 and 34) are in the same motion direction, but the heatmaps show that $D_{9 \rightarrow 16}$ is closer to $D_{9 \rightarrow 11}$ than $D_{9 \rightarrow 34}$. This result reveals that the transition from frame 9 to 16 is better among two potential candidates than to frame 34, i.e., there could be a noticeable jumping artifact in the transition from frame 9 to 34 in this context.} The average of $M_{\mathcal{D}}$ in this experiment is reported in Fig.\ref{fig_charts}-(a). The analysis shows that the stable rates of our rendered clips are relatively close to those in the source video. There are three cases (e.g., clip A, C, and F) in which the stable rates are relatively higher than the source. However, they are still at acceptable rates.

\subsubsection{\textbf{E.2. }\textcolor{black}{Degree of differences}} It is difficult to find a standard objective metric to measure \textcolor{black}{the differences of the generated sequences compared to the input ones}. Therefore, in this regard, we elaborate as follows.

We evaluate how different the rendered clips compared to the ground-truth by calculating the overlapping rate between them. To do this, we follow the well-known $\mathbf{F}$-measure \citep{mahasseni2017unsupervised} as the evaluation metric. \textcolor{black}{Previous works use this metric to measure the coherency of the rendered videos. The higher $\mathbf{F}$-measure is, the higher the coherent rate will be. Reversely, our purpose is to measure how different they are. To avoid confusion, we denote this value as $\Delta_o$. As a result, the smaller $\Delta_o$ represents the more difference.} Note that the clips generated by our system may be of different lengths and also less than those of the source clip. Let $G$ be the generated clip and $T$ be the corresponding source clip, the precision $P$ and recall $R$ is defined based on the amount of temporal overlap between $G$ and $T$, which are expressed as:
\begin{equation}
    P = \frac{\delta}{d_G}  \text{and } R = \frac{\delta}{d_T},
\end{equation}where $\delta$ is the duration of overlap between $G$ and $T$; $d_G$ and $d_T$ denotes the duration of clip $G$ and $T$, respectively. Finally, $\Delta_o$ is formulated as:
\begin{equation}
    \Delta_o = \frac{2 \times P \times R}{P + R} \times 100\%,
\end{equation}
Quantitative results on this aspect are shown in Fig.\ref{fig_charts}-(b). We can see that $\Delta_o$ of the testing data are relatively different from the ground-truth, especially on the clip D-Frog dance. It is worth pointing that the significant difference in this manner does not mean the clip is not stable. Inferring this clip in Fig.\ref{fig_charts}-(a), the results reveal that the sequence this clip is still stable. There are three cases (e.g., clip H, I, and J) where the different rate is low. This implies that these results are not significantly different with the ground-truth. The reason is that these cases consist of linear motion in the entire video. Therefore, our method can only generate the smooth sequence as the ground-truth in such cases. \textcolor{black}{In addition to these metrics, we conduct a user study to further learn about the human preferences on the visual quality of our results. Detail of the user study is described in the supplementary file.}

In summary, if we denote the total number of linear motion segment in a certain source video is $L$, the quality on these two aspects of the rendered clips is defined as follows. The stability ($M_D$) is covariate with $L$ and the differences of sequence ($\Delta_o$) is inverse with $L$. 

\subsubsection{\textbf{E.3. }\textcolor{black}{Human perception-based evaluation}} In addition to the above measurements, we further use human visual perception on the sequences generated by our method. Seven testing clips with small $\Delta_o$ are used in this evaluation. We first collect two \textcolor{black}{summarization} per sequence. Then we recruit a group of 11 users rank (in five levels) the \textcolor{black}{summarization} based on how well they describe the clip according to two questions. The detail of this study is described in the supplementary file.

For each question, let $s$ be the score if the $i^{th}$ user rates for the corresponding level of $s$ and $N_s$ be the number of rating of $s$. We use the following equation to compute the rate of each \textcolor{black}{summarization} to each question, which reflects the users' opinions:
\begin{equation}
    RA =  \bigg(\sum_{s=1}^{5} s \times N_s \bigg )/(5 \times 11) 
\end{equation}We then average $RA$ of two \textcolor{black}{summarizations} for each sequence to define the users' opinion. Fig.\ref{fig_charts}-(c) shows the statistics of users' preference. We can see that the scores of two questions are not extremely high but all of them are over the average degree (\textit{i.e.,} in range of 0.62 - 0.79, and 64\% is greater than 0.7). The results reveal that most users think the sequences generated by our work can tell meaningful stories.

\subsection{Comparisons to prior works}
We compare our system with some seminal works in this domain, including \citet{de2004cartoon}, \citet{yu2012combining}, \citet{3_yang2010recognizing}, and \citet{morace2022learning}. The different aspects in comparisons are summarized in Table \ref{table_compare}. In general, the early works \citep{de2004cartoon, yu2012combining, 3_yang2010recognizing} share the same two shortcomings: first, their mutual focus is the cartoon characters, and second, they need to do a pre-processing to extract the cartoon characters from the frames. Manifold method \citep{morace2022learning} is more general, i.e., it does not need such a pre-processing and thus, it is adaptable to cartoon scenes. However, they do not consider the motion direction as the other competitors \citep{de2004cartoon, yu2012combining, 3_yang2010recognizing} do, clips with dense motion directions are the major limitation in their system. \textcolor{black}{In shape contrast}, our approach has three major advantages. First, our system performs well on arbitrary input video scenes. Second, our system does not need any pre-processing. And third, our system is able to produce novel animations compared with those in the given video. The remainder of this subsection describes detailed comparison on each single competitor.

Fig.\ref{fig_compare_daffy} shows a qualitative comparison between our results and those in \citet{de2004cartoon}. The pair of frames in (b) is mentioned as a bad transition in \citep{de2004cartoon}. As a result, they have to insert inbetweens to obtain good transition. In contrast, our method automatically defines the adjacent frame with a smooth transition without refinement. It is observed that our transition in (a) is more plausible compared to (b). 

Similar to our approach, \citet{3_yang2010recognizing} consider motion direction in transitions. The significant difference here is that they focus on cartoon characters. Gesture of characters needs to be extracted to define the similarity between frames (\textcolor{black}{see Fig.7 in the supplementary file}). Moreover, the motion direction feature (MDF) cannot accurately describe the gesture of a cartoon character. Thus, their approach is not effective to explore the challenging input. Reversely, our system gets knowledge from self-defined network to learn the similarity of images in terms of content correlation and embeds optical flow to maintain consistency in motion directions. Therefore, we advance not only in arbitrary input but also in accuracy.

Fig.\ref{fig_compare_fish} shows the comparison with \citet{morace2022learning}. The source clip of this example consists of dense motions of fish and chinese ink, in which there exist several linear motion segments. As in our early discussion, since \citet{morace2022learning} do not consider the motion direction, there is significant abrupt motion in the regions masked in red rectangles. Fortunately, thanks to the constraints in our scheme, we resolve this phenomenon and obtain smooth transitions in the generated sequence. Another aspect makes \citep{morace2022learning}'s system suffer some limitations (\textit{i.e.}, image content is complex) is that they use LPIPS metric to define the similarity of image pairs. This metric is learned by training a \say{small network} which is designed to predict perceptual judgement from distance pair and not originally designed for resequencing application. Besides, it takes approximately 5 seconds to compute on a pair. Therefore, the performance of \citep{morace2022learning} heavily relies on those in this model.

\input{Compare}
\input{Compare_Daffy}
\input{Compare_fish}
\input{Ablation_3_RFF}

\textcolor{black}{Apart from the above visual comparisons, we quantitatively compare the quality of our results against those of prior work by two metrics $M_D$ and $\Delta_o$. We also use the data in Fig.\ref{fig_testing_set} in this comparison. And our competitor is Manifold \citep{morace2022learning} since the other three methods \citep{de2004cartoon, 3_yang2010recognizing, yu2012combining} focus on cartoon characters, and their results are not available for a fair comparison. Meanwhile, Manifold \citep{morace2022learning}’s focus is comparable to ours, and the source code is provided by the authors. Table \ref{table_compare_manifold} presents the statistic results in this comparison. We can see that our method outperforms on the average of stability score. In terms of $\Delta_o$, Manifold and ours have the comparable scores. However, we can see that their values of $\Delta_o$ are relatively equal, and the score in cartoon data (A-G) are higher than natural scenes (H-K). When we inspect $M_D$ of data A-G, they are not at good stability degree. This reveals that \citet{morace2022learning} fail to either generate new sequences for cartoon data or produce smooth sequences with linear motion in natural scene data. Conversely, in our method, smaller $\Delta_o$ on cartoon data implies that it can explore new sequences. For the natural scene data with linear motion, higher $\Delta_o$ side by side with smaller $M_D$ reveal that it can tolerate to avoid flip-flop phenomenon in such data.}

\subsection{Ablation Study}  
\subsubsection{\textcolor{black}{\textbf{A.1. }Verify the effectiveness of RSFNet}}
Our RSFNet is structured in the integration of a backbone and the proposed RFF module. Without RFF module, generated sequences include inconsistencies due to the lack of information on the features that are extracted from the backbone. We demonstrate the effectiveness of RFF module by removing it from our training. We show these ablation analyses in Fig.\ref{fig_Ablation_3}. Here, we visualize the Grad-CAMs \citep{selvaraju2017grad} of those obtained from our RSFNet with and without RFF module. The results show that with FRR module, our RSFNet has much larger attended regions. This enables our system to have more predictable results.

\subsubsection{\textcolor{black}{\textbf{A.2. }Study on the impact of distance metric}}
\textcolor{black}{Performance of our resequencing system is affected by the feature correlation calculation. To analyze the influence of feature correlation on the quality of rendered sequences, we change the model to calculate the distance metric by a pure Euclidean distance calculation. That is, we remove the distance loss (\textit{e.g.}, Eq. (4)) and use the Euclidean distance to measure the correlation in pairs of latent vectors. Fig.\ref{fig_distance_metric} shows the contrast results. It is observed that Euclidean distance metric performs the correlation of the neighbors well. For example, we inspect on frame $5$, which is highlighted in green rectangle. We can see that most similar frames are adjacent frames of this frame (\textit{e.g.}, frame 4, 5, 7). Meanwhile, our distance is able to capture more (\textit{e.g.}, frame 1, 2, 3, 4, 5, 12, 20}). Therefore, if we directly use Euclidean as the distance metric, it prevents us from exploring \textcolor{black}{new} paths. 

\subsubsection{\textbf{A.3. }Study on constraints in SDPF}
\textbf{A.3.1. Directional distillation. }\textcolor{black}{This constraint is configured to detect the motion's property of a certain frame. As we mentioned in previous session, the \say{property} here is the linear motion. To verify the impact of this constraint ($C_d$) in the results, we remove it from the full procedure. That is, $C_d$ is omitted from Eq.(18). Fig.\ref{fig_ablated_Cd} shows the results of ablation analysis. In this example, we deliberately choose a frame (\textit{i.e.}, frame 187) that belongs to such a linear motion segment to clearly reveal the influence of this constraint. After the first distillation, we define five candidates that have feature correlation to frame 187. Among them, frame 152 does not belong to LMS, meanwhile, the remainders are. In the remained candidates, frame 121 and 144 are in reverse direction motion with frame 187, and frame 193 is the same direction with frame 187. Without constraint $C_d$, frame 152 and 135 are selected as being adjacent with frame 187. Obviously, the flip-flop phenomenon will occur. Reversely, with constraint $C_d$, frame 193 is chosen. This result yields a reasonable transition. } 

\input{vs_manifold}
\input{distance_metric}
\input{Ablated_C_d}

\textbf{A.3.2. Coherent distillation. }\textcolor{black}{Without this constraint, the motion in generated sequences can be realistic but may fail in temporal coherence. We measure this effect quantitatively by removing this constraint (i.e., $C_t$) from our proposed procedure. Fig.\ref{fig_ablated_Ct} visualizes the ablated results in this aspect. It is observed that without $C_t$, the adjacent frame of frame 143 is frame 172. In the contrast case, it is frame 149. Although both frame 149 and 172 are the same direction motion with frame 143, the heat maps reveal that }the differences from frame 143 to 172 is significant. This is the reason that causes the jumping transition in the rendered clips without $C_t$.

\input{Ablated_Ct}

\input{Ablated_constraints}

\textcolor{black}{In summary, we verify the effectiveness of RSFNet, distance loss, and two constraints ($C_d$, $C_t$) by testing on 12 videos in Fig.\ref{fig_testing_set}. The analysis is shown in Table \ref{table_ablated_constraints}. From these results, we can conclude that the each constraint plays an important role for the stability of the rendered clips; the distance metric and RSFNet affect to the ability in exploring new sequences. Full configuration guarantees better quality results. }

\subsection{Limitations}
In the cases that the input videos consist of subtle motion of landscape scenes (see the visualized sample here\footnote{\href{http://graphics.csie.ncku.edu.tw/SDPF/Failure.mp4}{http://graphics.csie.ncku.edu.tw/SDPF/Failure.mp4}}), our method may not perform well. The failure phenomenon in such data is that the resultant sequence is quite short, i.e., approximately 20\% of the total number of frames in the source video. We note here that these results are still smooth. The reason is our SDPF utilizes the temporal coherency or the velocity of motion in the source video to estimate the adjacent frame in each single pair of frames. In such subtle motion, the differences of the adjacent frames are small and the motion is looped. Therefore, our SDPF will early stop if the changes are relatively large to avoid cold transitions.

\section{Conclusion}
In this paper, we propose a new RETVI framework for retargeting videos. With two modules configured in our method, our RETVI presents high performance in handling videos with diverse contents and produces visually pleasing results when retargeting to arbitrary aspect ratios. The analysis and experimental results demonstrate that our method substantially advances prior works. With the fast running time of our end-to-end RETVI, our system is potentially embedded into a video resizing application/service. \textcolor{black}{We perceive that our system can bypass the computational bottlenecks in conventional methods. And it is potential to extend for stereo image/video retargeting}. For the shortcoming we discussed, we plan to investigate techniques that configure the loss function to be independent from the existing feature extractor. \textcolor{black}{In terms of cropping effect, a possible way can improve is automatically define physical region of the important content. This knowledge could serve us to shift the rendering window more appropriately}. Besides, developing a text-driven framework to consider semantic issue in the retargeting videos \textcolor{black}{and investigating technique to retarget videos with enlarging and reducing two dimensions simultaneously are also a possibly extension in our near future}. This could be a potential way to visualize users' expectation in such a video retargeting system.

\bibliographystyle{abbrv}
\bibliography{references}
\end{document}

%% file: overall_framework.tex
\begin{figure}[h!]
  \centering
  \includegraphics[width=0.75\linewidth]{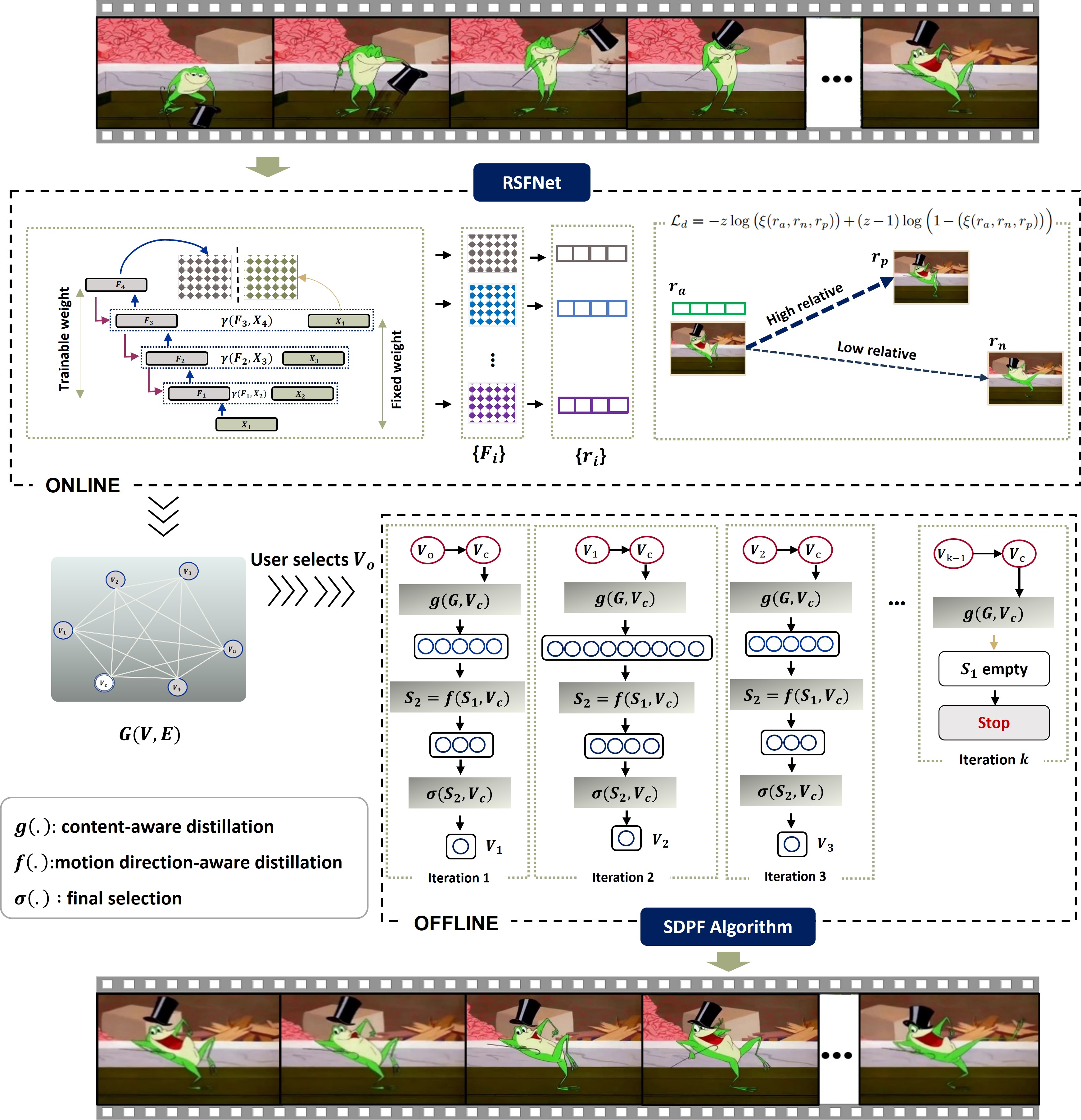}
  \caption{\small Our framework for regenerating video sequence.}
  \label{fig_framework}
\end{figure}

%% file: Encoder.tex
\begin{figure}[h!]
  \centering
  \includegraphics[width=\linewidth]{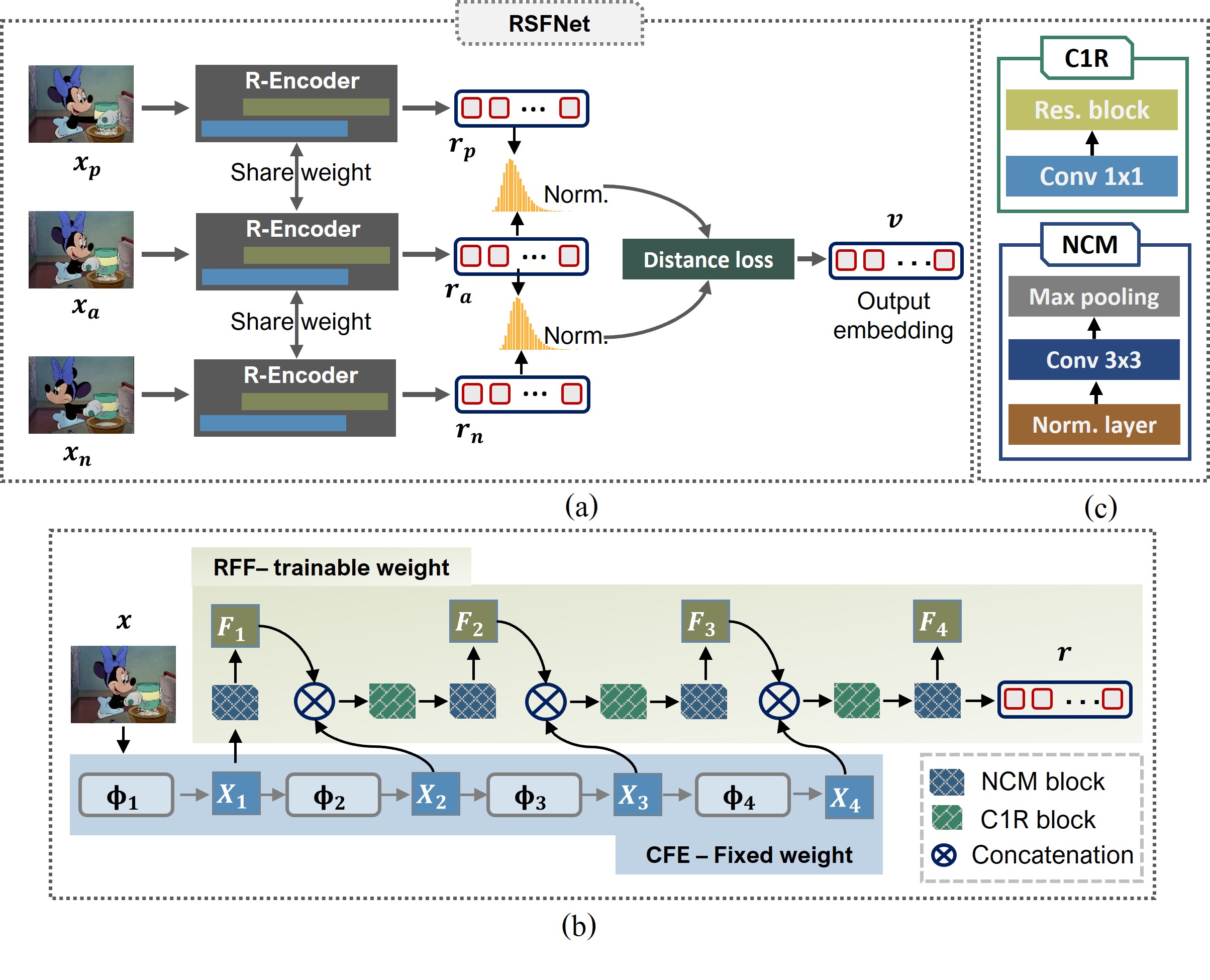}
  \caption{\textcolor{black}{\small (a) Architecture of RSFNet; (b) Zoom-in of an R-Encoder; (c) Structure of NCM and C1R block.}} 
  \label{fig_encoder}
\end{figure}

%% file: LMS_visualize.tex
\begin{figure}[h]
  \centering
  \includegraphics[width=0.8\linewidth]{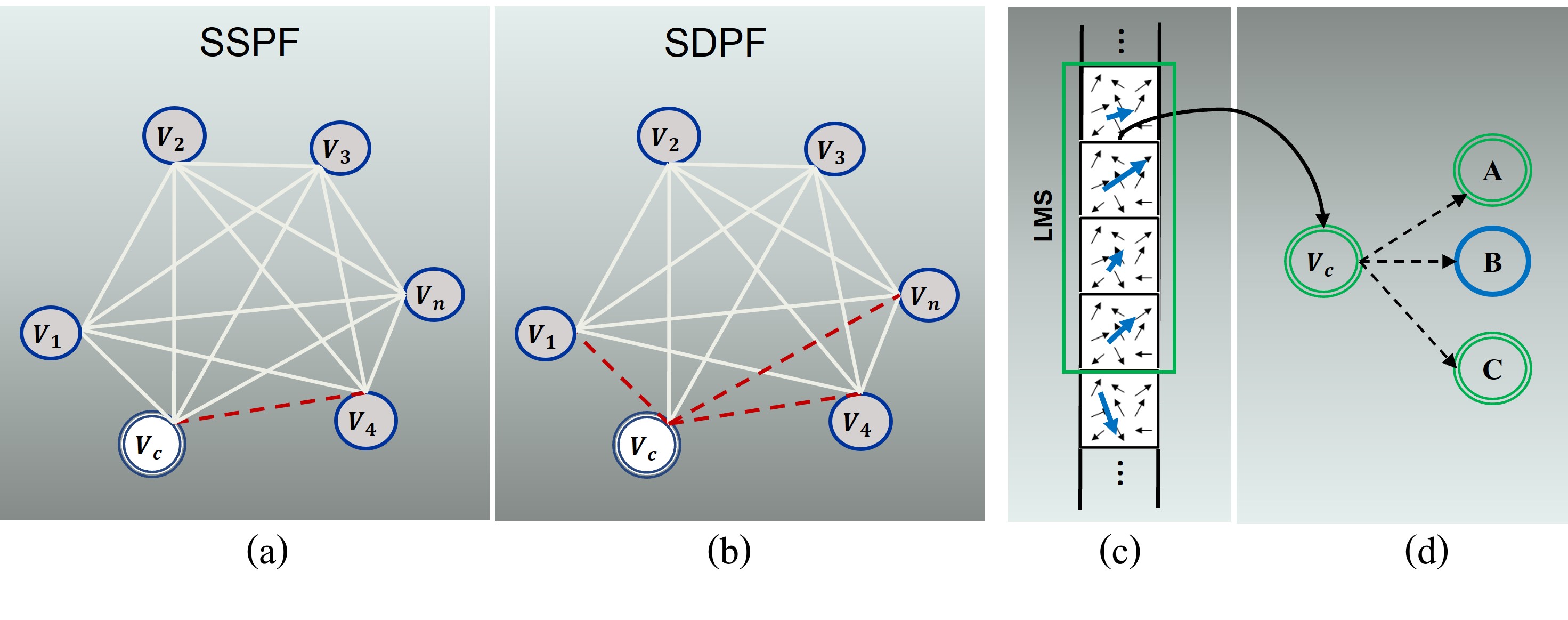}
  \vspace{-0.3cm}
  \caption{\small Visualization of SSPF (a) and SDPF (b). (c) is visualization of motion tendency of each frame (\textit{i.e.}, blue arrow). The frames in green rectangle belong to an LMS, \textit{i.e., }the adjacent frames satisfy Eq.(11). (d) is visual constraint $C_d$ in Eq.(12), here the node outlined by double circle means it is an LMS-frame.}
  \label{fig_LSM_visualize}
\end{figure}

%% file: PMSM.tex
\begin{figure}[h]
  \centering
  \includegraphics[width=0.9\linewidth]{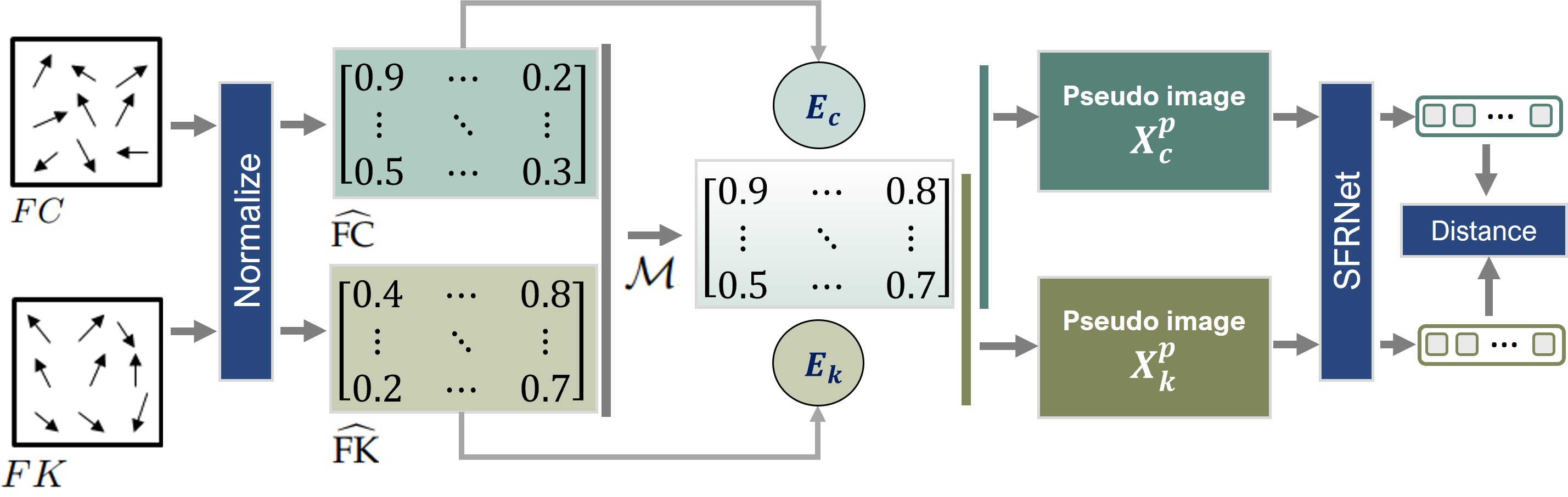}
  \caption{\textcolor{black}{\small Flowchart of our proposed PMSM}.}
  \label{fig_DOF}
\end{figure}

%% file: Testing_set.tex
\begin{figure}[h]
  \centering
  \includegraphics[width=\linewidth]{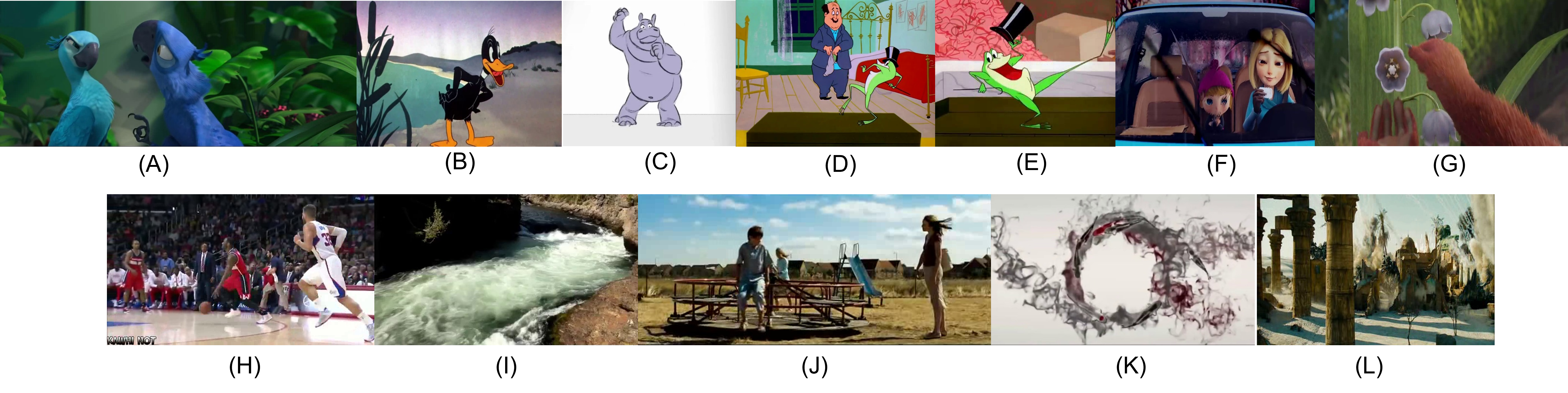}
  \caption{\small Frames in some of the videos we use to evaluate our method.}
  \label{fig_testing_set}
\end{figure}

%% file: different_sequences.tex
\begin{figure}[h]
  \centering
  \includegraphics[width=\linewidth]{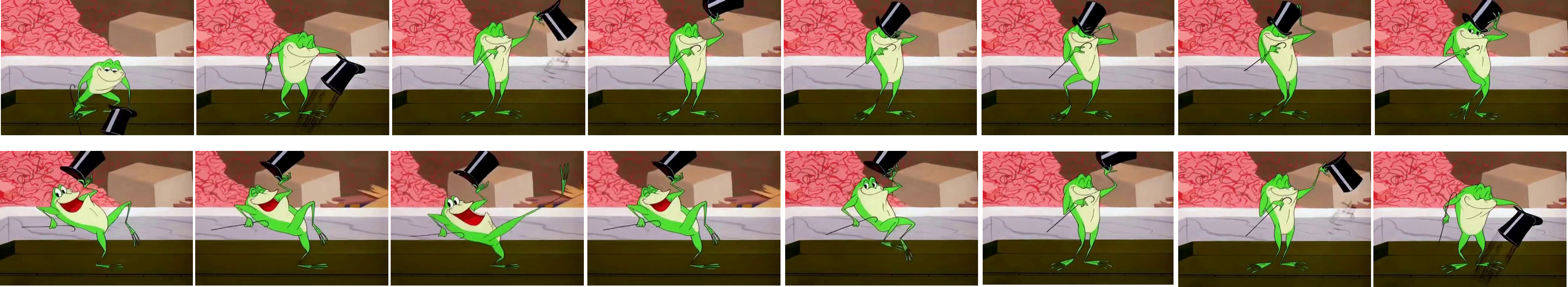}
  \caption{\small Demonstrates the differences in the sequence generated by our method versus those in the original video. Shown in this figure are the filmstrips from original video (first row) and our rendered video (second row).}
  \label{fig_different_sequences}
\end{figure}

%% file: Differences_Frames.tex
\begin{figure}[h]
  \centering
  \includegraphics[width=\linewidth]{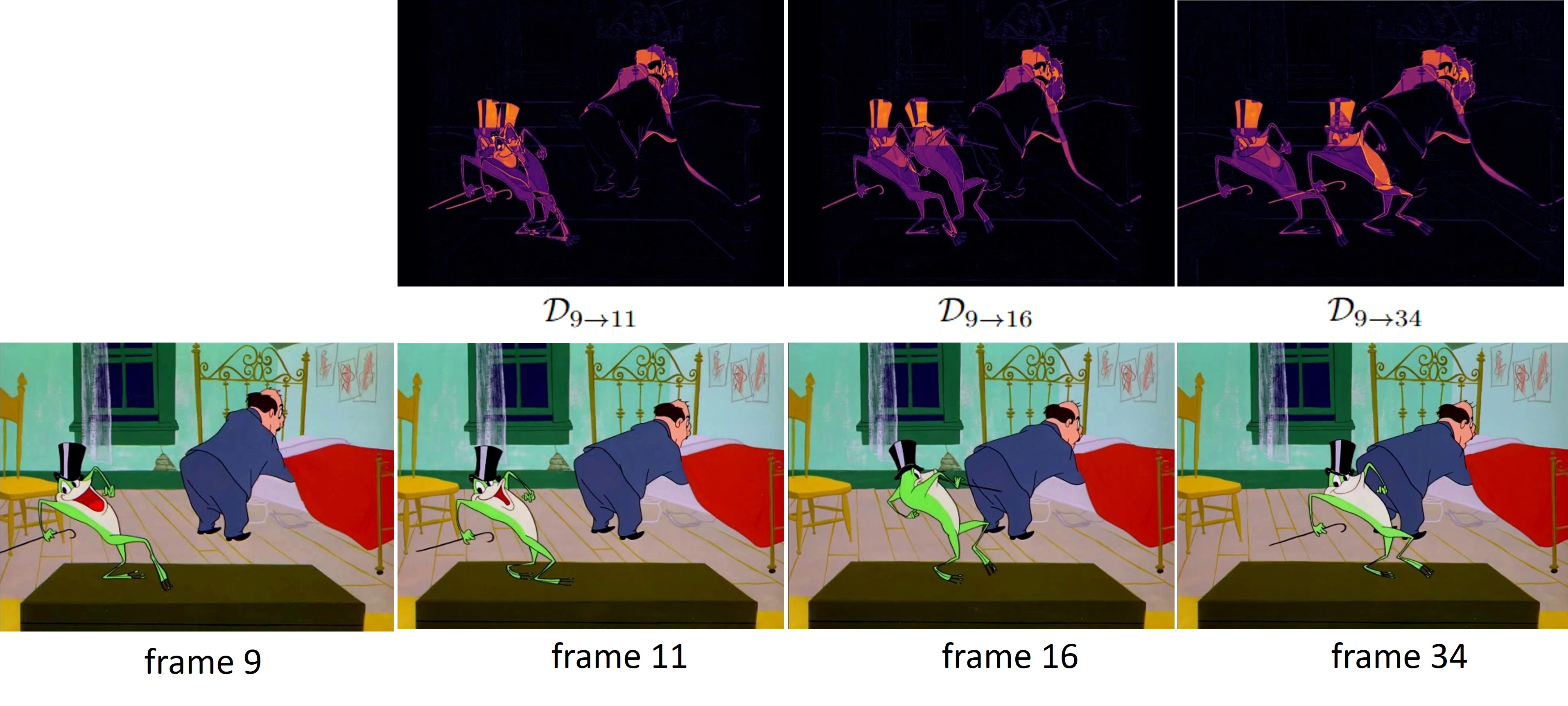}
  \caption{\small Visualizes the heatmap of differences of frames.}
  \label{fig_differences_frames}
\end{figure}

%% file: Charts.tex
\begin{figure*}[h]
  \centering
  \includegraphics[width=\linewidth]{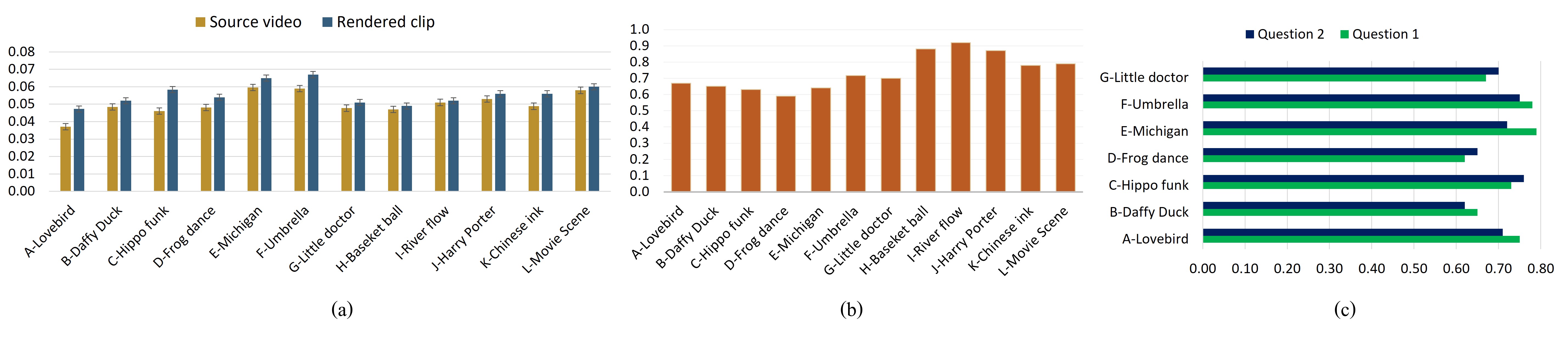}
  \vspace{-0.3cm}
  \caption{\small Analysis results on (a) stability, (b) difference degree, and (c) human perception on our resequencing results.}
  \label{fig_charts}
\end{figure*}

%% file: Compare.tex
\begin{table}[h]
\caption{\centering \small Comparisons between our method and prior works} 
\centering 
\begin{tabular}{l c c c} 
\\ [0.5ex] 
 \Xhline{2\arrayrulewidth} 
Methods &  Pre-processing &  New sequence? & Type of data \\ [0.5ex] 
\cmidrule(lr){1-4}
GCCS \citep{de2004cartoon} & Yes & No & Cartoon characters\\ [1ex] 
RCCS \citep{3_yang2010recognizing} & Yes & No & Cartoon characters\\ [1ex] 
semi-MSL \citep{yu2012combining} & Yes & No & Cartoon characters\\  [1ex] 
Manifold \citep{morace2022learning} & No & No & Cartoon scenes\\  [1ex] 
Our method  & No & Yes & Arbitrary scenes\\
\Xhline{2\arrayrulewidth} 
\end{tabular}
\label{table_compare} 
\end{table}

%% file: Compare_Daffy.tex
\begin{figure}[h]
  \centering
  \includegraphics[width=\linewidth]{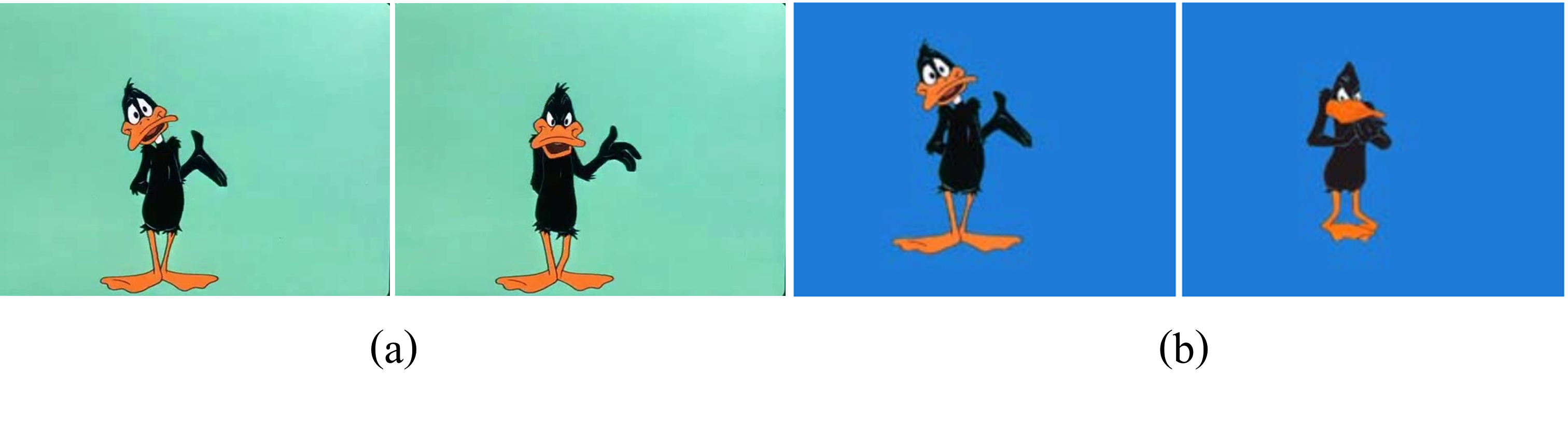}
  \vspace{-0.3cm}
  \caption{\small Visualizes the differences in transition between our method (a) and \citep{de2004cartoon} (b). Photos in (b) are obtained from \citep{de2004cartoon}.}
  \label{fig_compare_daffy}
\end{figure}

%% file: Compare_fish.tex
\begin{figure}[h]
  \centering
  \includegraphics[width=\linewidth]{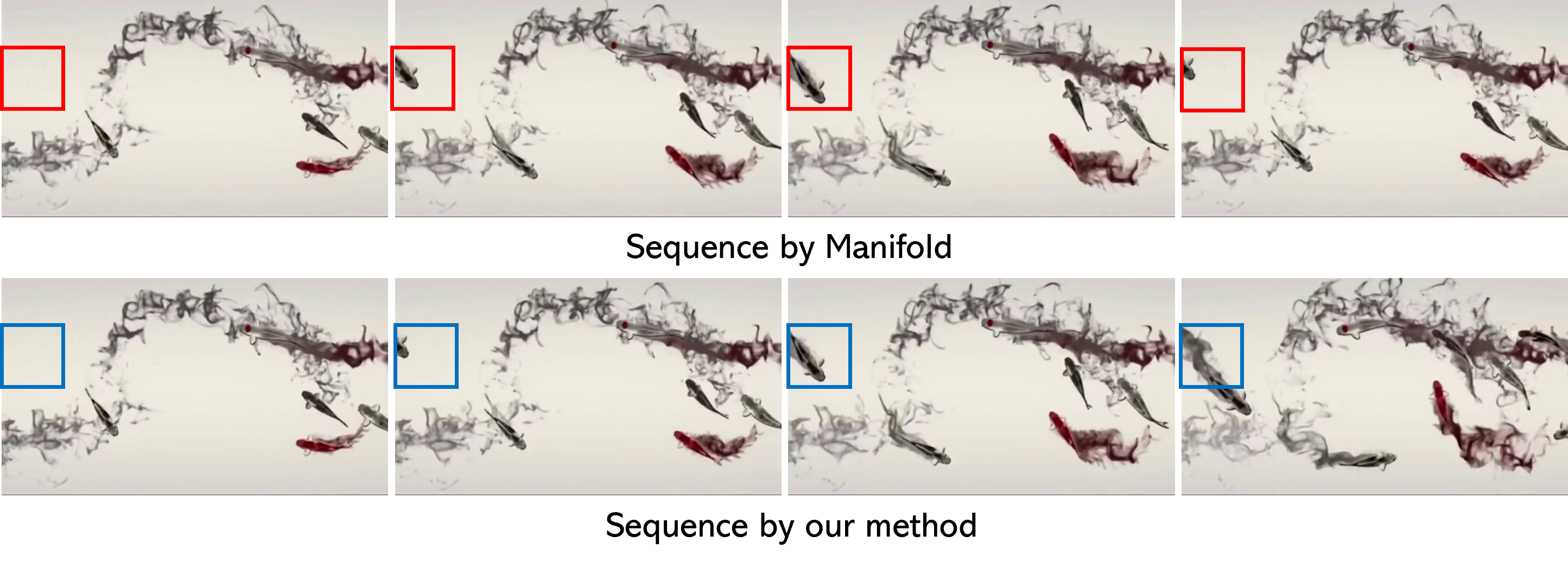}
  \vspace{-0.2cm}
  \caption{\small Comparisons with Manifold sequence \citep{morace2022learning}.}
  \label{fig_compare_fish}
\end{figure}

%% file: Ablation_3_RFF.tex
\begin{figure}[h]
  \centering
  \includegraphics[width=\linewidth]{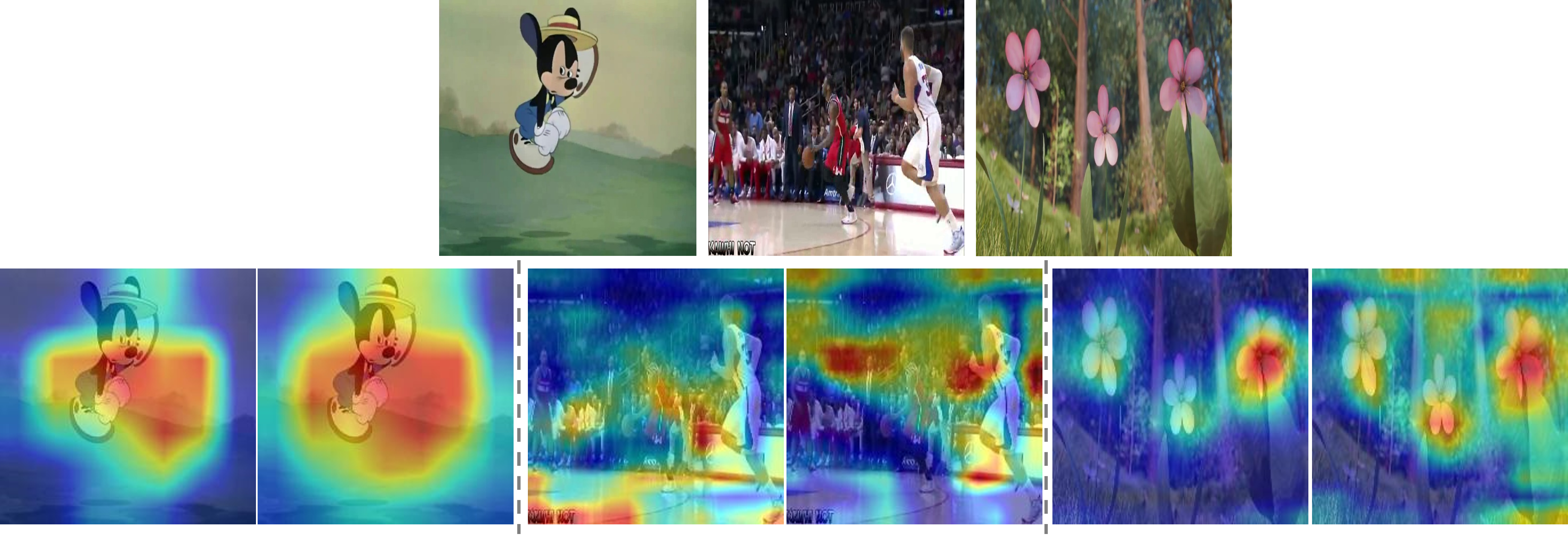}
  \caption{\small Second row is the Grad-CAM visualization of the backbone VGG-19 (left) and our RSFNet (right).}
  \label{fig_Ablation_3}
\end{figure}

%% file: vs_manifold.tex
\begin{table}[h!]
  \centering
  \caption{\textcolor{black}{\small Comparisons on the quality of results }}
    \begin{tabular}{lccccc}\\
    \Xhline{2\arrayrulewidth} 
          Methods& Ground-truth & \multicolumn{2}{c}{\citet{morace2022learning}} & \multicolumn{2}{c}{Our method} \\
          \cmidrule(lr){1-1}\cmidrule(lr){2-2} \cmidrule(lr){3-4} \cmidrule(lr){5-6}
          Testing data & \textcolor{black}{$M_D$}   & \textcolor{black}{$M_D$}   & $\Delta_o$ & \textcolor{black}{$M_D$}   & $\Delta_o$ \\
        \cmidrule(lr){1-6}
        A- Lovebird & \textcolor{black}{0.037}      & \textcolor{black}{0.056}    & 0.78    &
        \textcolor{black}{0.047}    & 0.67  \\ [1ex] 
        B- Daffy Duck & \textcolor{black}{0.048}       & \textcolor{black}{0.067}    & 0.84    & \textcolor{black}{0.052}    & 0.64  \\[1ex] 
        C- Hippo funk & \textcolor{black}{0.046}       & \textcolor{black}{0.062}    & 0.81    & \textcolor{black}{0.058}    & 0.63  \\[1ex] 
        D- Frog dance & \textcolor{black}{0.048}       & \textcolor{black}{0.079}    & 0.77    & \textcolor{black}{0.054}    & 0.59  \\[1ex] 
        E- Michigan & \textcolor{black}{0.060}       & \textcolor{black}{0.062}    & 0.79    & \textcolor{black}{0.065}    & 0.64  \\[1ex] 
        F- Umbrella & \textcolor{black}{0.059}      & \textcolor{black}{0.083}    & 0.69    & 
        \textcolor{black}{0.067}    & 0.72  \\[1ex] 
        G- Little doctor & \textcolor{black}{0.048}      & \textcolor{black}{0.073}    & 0.73    & \textcolor{black}{0.051}    & 0.70  \\[1ex] 
        H- Basket ball & \textcolor{black}{0.047}      & \textcolor{black}{0.081}    & 0.58    & \textcolor{black}{0.049}    & 0.88  \\[1ex] 
        I- River flow & \textcolor{black}{0.051}      & \textcolor{black}{0.075}    & 0.75    & \textcolor{black}{0.052}    & 0.92  \\[1ex] 
        J- Harry Porter & \textcolor{black}{0.053}        & \textcolor{black}{0.078}    & 0.68    & \textcolor{black}{0.056}    & 0.87  \\[1ex] 
        K- Chinese ink & \textcolor{black}{0.049}      & \textcolor{black}{0.052}    & 0.73    & \textcolor{black}{0.056}    & 0.78  \\[1ex] 
        L- Movie Scene & \textcolor{black}{0.058}       & \textcolor{black}{0.065}    & 0.74    & \textcolor{black}{0.060}    & 0.79  \\[1ex] 
        \cmidrule(lr){1-6}
        \textbf{Average} & \textcolor{black}{0.050}       & \textcolor{black}{0.069}    & 0.74    & \textcolor{black}{0.055}    & 0.73  \\
    \Xhline{2\arrayrulewidth} 
    \end{tabular}%
  \label{table_compare_manifold}%
\end{table}%

%% file: distance_metric.tex
\begin{figure}[h!]
  \centering
  \includegraphics[width=0.8\linewidth]{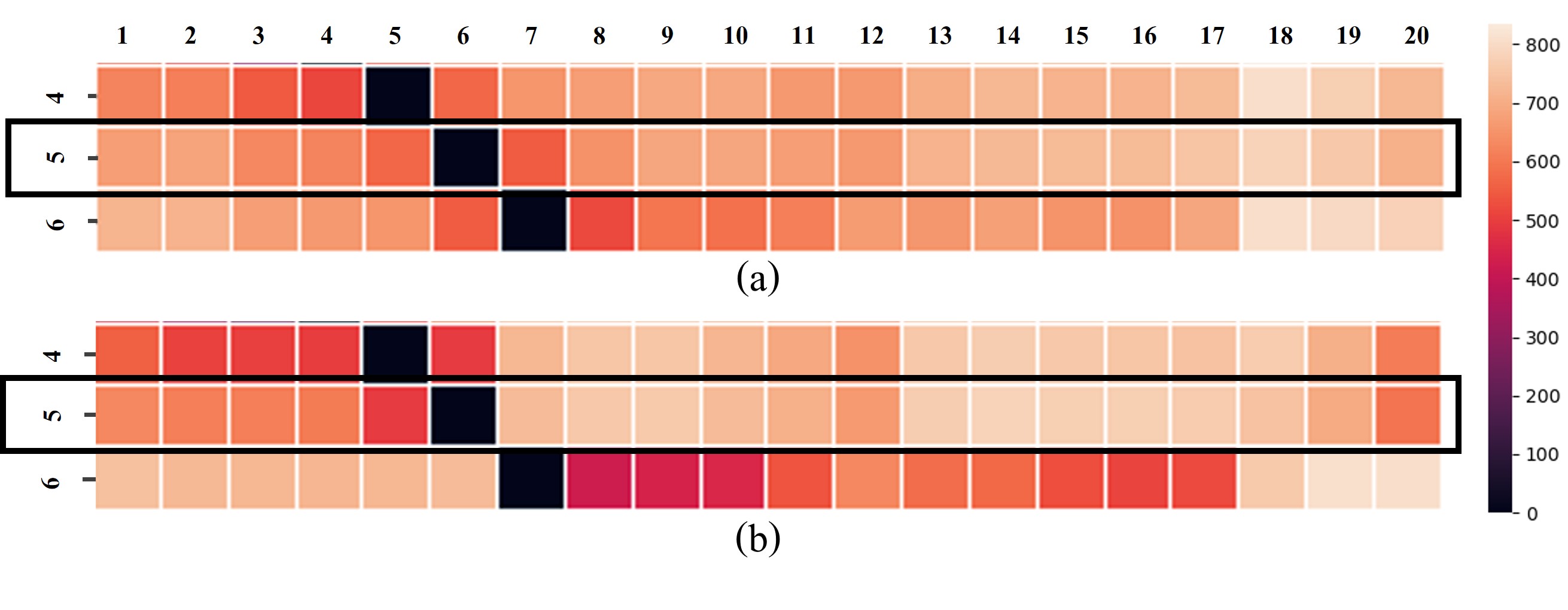}
  \vspace{-0.3cm}
  \caption{\textcolor{black}{\small Zoom-in the heat map of distance metric calculated by Euclidean distance (a) and our learning-based Euclidean distance (b). The experiment is conducted on segment with 20 frames of Daffy Duck clips. Entire heat maps could be seen in Fig.6 in the supplementary file.}}
  \label{fig_distance_metric}
\end{figure}

%% file: Ablated_C_d.tex
\begin{figure}[h!]
  \centering
  \includegraphics[width=\linewidth]{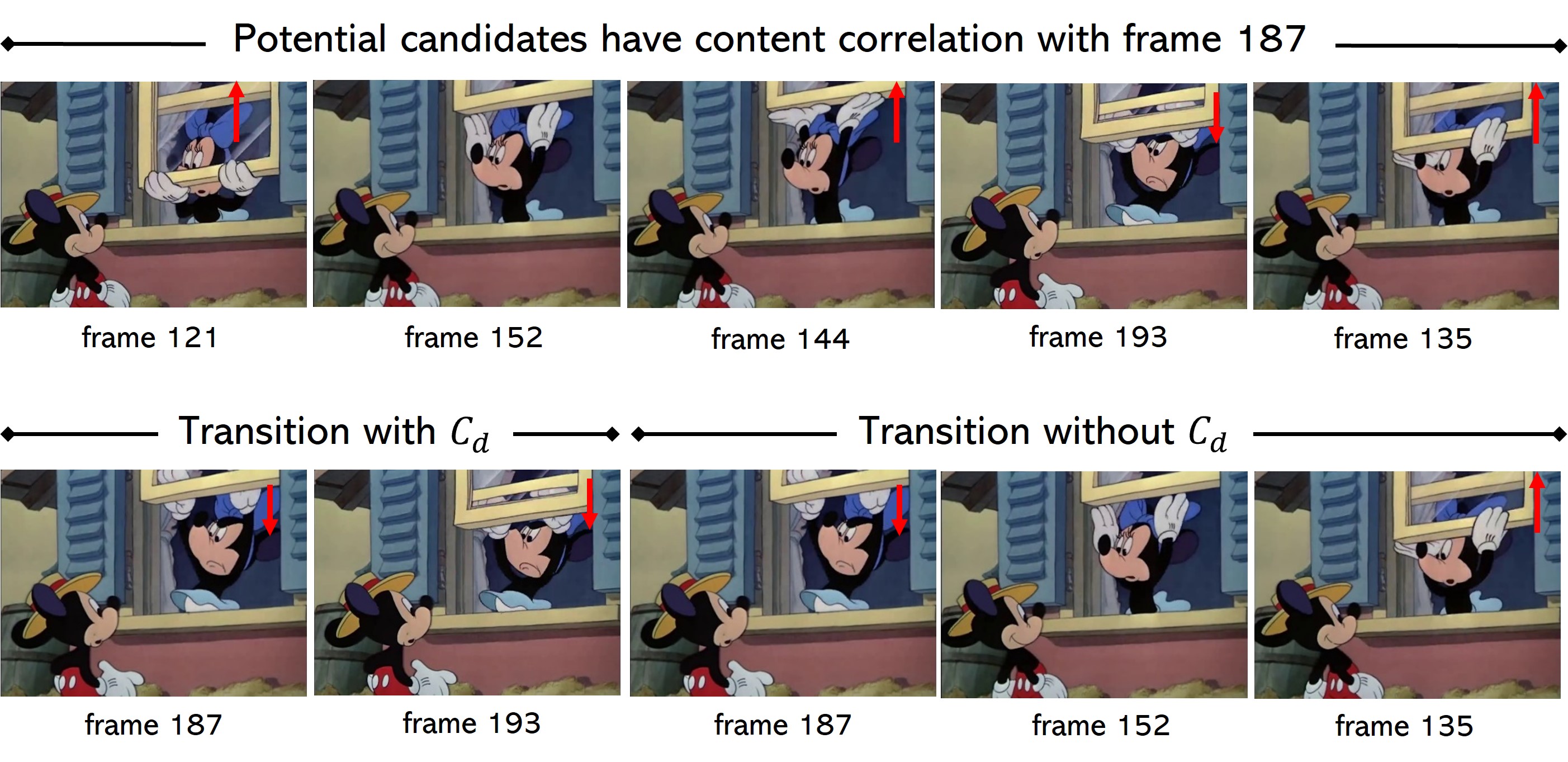}
  \caption{\small Ablated results of constraint $C_d$. Five candidates in $\mathcal{S}_1$ are correlated with frame 187. Because frame 187 belongs to LMS, if $\mathcal{S}_1$ has LMS-frames, they will be considered to guarantee the coherency with frame 187. With $\mathbf{C}_t$, frame 193 is chosen. This is an LMS-frame, we can see the transition is visual smooth. In the contrast case, both frame 152 and 135 do not belong to LMS, exploring to these frames may cause artifacts.}
  \label{fig_ablated_Cd}
\end{figure}

%% file: Ablated_Ct.tex
\begin{figure}[h]
  \centering
  \includegraphics[width=\linewidth]{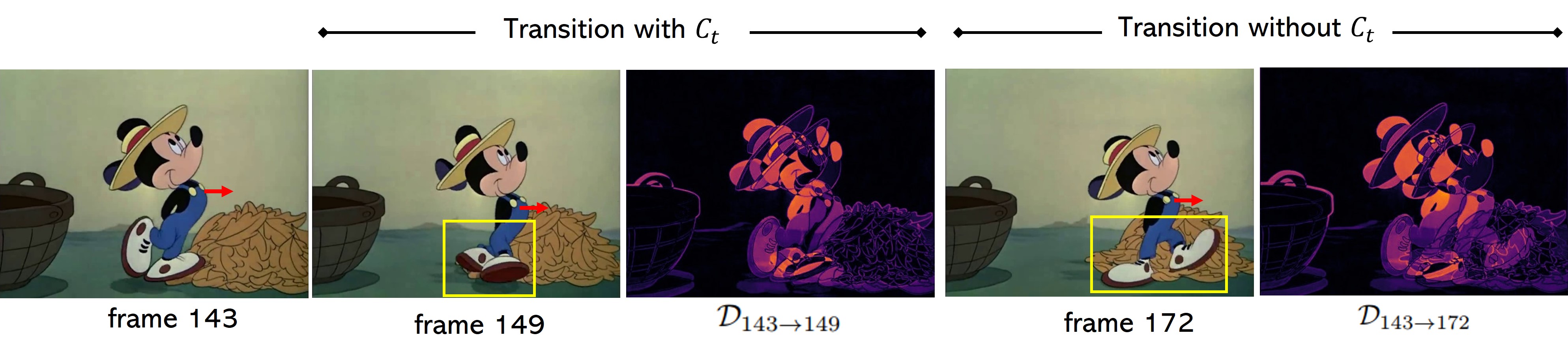}
  \caption{\small Ablated results of constraint $C_t$}
  \label{fig_ablated_Ct}
\end{figure}

%% file: Ablated_constraints.tex
\begin{table}[H]
\caption{\centering \textcolor{black}{\small Analysis of the quality on ablated results } }
\centering 
\begin{tabular}{l c c } \\
\Xhline{2\arrayrulewidth} 
 Method & \ $M_D$ & \ $\Delta_o$ \\ [0.5ex] 
\cmidrule(lr){1-3}
Ground-truth & 0.0535 & 1 \\ [1ex] 
w/o $C_t$ & 0.118 & 0.783 \\ [1ex] 
w/o $C_d$  & 0.079 & 0.827 \\  [1ex]
w/o RSFNet  & 0.082 & 0.731 \\   [1ex]
w/o $\mathcal{L}_d$  & 0.281 & 0.816 \\  [1ex] 
Full configure & 0.0648 & 0.706 \\
\Xhline{2\arrayrulewidth} 
\end{tabular}
\label{table_ablated_constraints} 
\end{table}